\documentclass[arxiv]{melba}

\usepackage{mwe} 

\usepackage{amsmath,amsfonts}
\usepackage{graphicx}

\usepackage[hidelinks]{hyperref}
\hypersetup{ colorlinks, 
            citecolor=black, 
            filecolor=black, 
            linkcolor=black, 
            urlcolor=black}
\usepackage{rotating}
\usepackage[export]{adjustbox}
\usepackage{caption}
\usepackage{subcaption}
\usepackage[section]{placeins}
\usepackage{threeparttable}
\usepackage[T1]{fontenc}

\melbaid{YYYY:NNN}  
\doi{https://doi.org/10.59275/j.melba.2024-AAAA}
\melbaauthors{Wiedeman and Sarmakeeva}  
\volume{2}
\firstpageno{1}  
\melbayear{2025}  
\datesubmitted{m1/yyyy}  
\datepublished{m2/yyyy}  

\melbaspecialissue{Medical Imaging with Deep Learning (MIDL) 2020}
\melbaspecialissueeditors{Marleen de Bruijne, Tal Arbel, Ismail Ben Ayed, Hervé Lombaert}

\ShortHeadings{T-SYNTH: A Knowledge-Based Dataset of Synthetic Breast Images}{Wiedeman and Sarmakeeva}

\title{T-SYNTH: A Knowledge-Based Dataset of Synthetic Breast Images}

\author{
Christopher Wiedeman$^*$, Anastasiia Sarmakeeva$^*$, Elena Sizikova, Daniil Filienko, Miguel Lago, Jana G. Delfino, \\ Aldo Badano}

\affiliations{Office of Science and Engineering Laboratories\\ 
Center for Devices and Radiological Health\\ 
U.S. Food and Drug Administration\\ 
Silver Spring, MD 20993 USA \\
$*$ - These authors contributed equally to this work.}

\abstract{
One of the key impediments for developing and assessing robust medical imaging algorithms is limited access to large-scale datasets with suitable annotations. Synthetic data generated with plausible physical and biological constraints may address some of these data limitations. We propose the use of physics simulations to generate synthetic images with pixel-level segmentation annotations, which are notoriously difficult to obtain. Specifically, we apply this approach to breast imaging analysis and release T-SYNTH, a large-scale open-source dataset of paired 2D digital mammography (DM) and 3D digital breast tomosynthesis (DBT) images. Our initial experimental results indicate that T-SYNTH images show promise for augmenting limited real patient datasets for detection tasks in DM and DBT. Our data and code are publicly available at~\url{https://github.com/DIDSR/tsynth-release}.}


\begin{document}

\twocolumn[\maketitle]

\section{Background}
\label{sec:intro}
Responsible for approximately two million new cases and over six hundred thousand deaths in 2022 alone~\citep{sung2020global}, breast cancer remains a prominent global health concern, and is expected to account nearly one-third of all newly diagnosed cancers among women in the United States~\citep{desantis2016breast}. According to the most recent report from International Agency for Research on Cancer~\citep{bray2022global}, it is one of the most widespread cancers diagnosed worldwide, both in the number of cases and associated deaths. Consequently, medical imaging techniques are indispensable for screening, diagnosis, and further research into the disease. Historically, the most common imaging technique for breast cancer screening is digital mammography (DM), in which a 2D x-ray projection of a compressed breast is taken. Digital breast tomosynthesis (DBT), a pseudo-3D imaging technique, has been increasingly adopted, demonstrating improved screening performance~\citep{asbeutah2019comparison, sprague2023digital}.

\begin{figure*}[htb]
    \centering
    \includegraphics[width=0.9\linewidth]{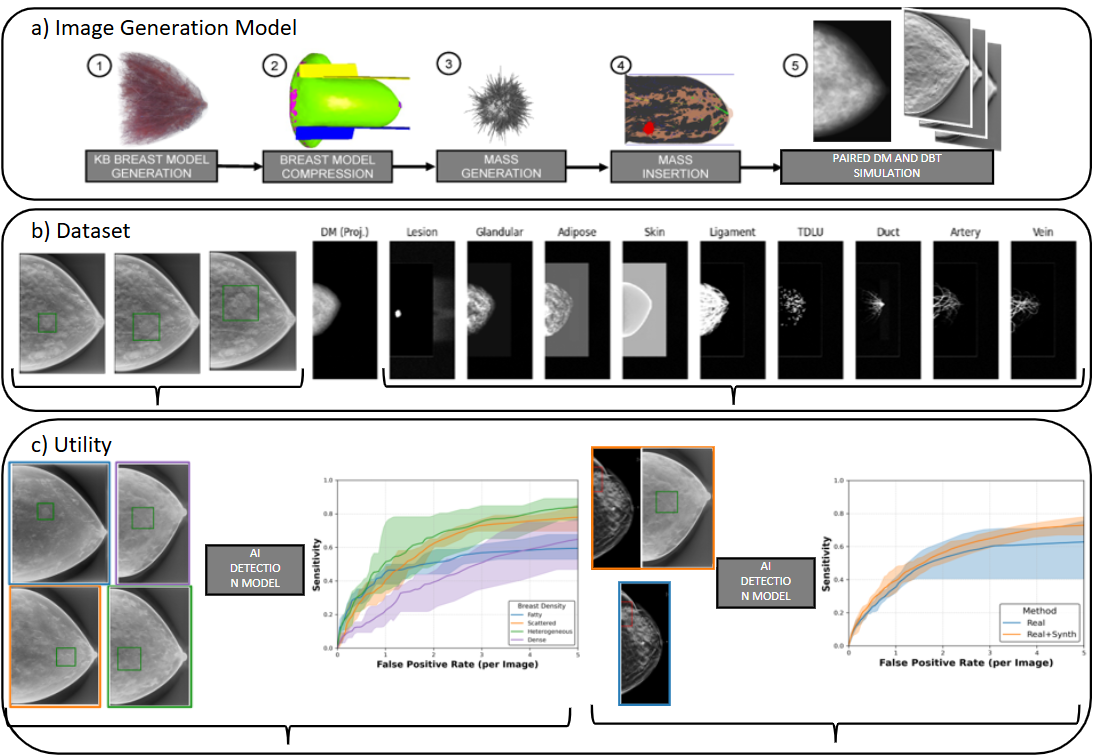}
    \caption{Summary of T-SYNTH: a) Knowledge-based process for simulating paired synthetic digital mammography (DM) and digital breast tomosynthesis (DBT) examples. b) Synthetic T-SYNTH data, which includes varying lesion sizes, lesion densities (left), and tissue masks (right). c) Utility of T-SYNTH, which includes comparing performance of AI models for lesion detection in different subgroups (left) and augmenting existing training data with T-SYNTH for better detection in patient breast images (right).}
    \label{fig:summary_figure}
\end{figure*}

The development of algorithms for segmenting, detecting, or classifying structures of interest (e.g., cancerous lesions) within these images is desirable for improving both patient outcomes as well as workflow efficiency. A limiting factor to developing data-driven algorithms, however, is the access to representative datasets with appropriate annotations. Collecting representative medical imaging data is challenging due to cost, patient privacy regulations, and other constraints, but obtaining appropriate annotations is even more taxing, since annotators must have specialized domain knowledge. The labeling process itself is laborious due to the high resolution and fine details typically present in these images. Very few DM and DBT datasets are publicly available. Furthermore, the cohort characteristics and amount of information included in these datasets are highly variable, which limits the robustness of their application. Only a fraction of publicly-available datasets contain pixel-level annotations for relevant tissues, since pixel-level annotations typically require a time-consuming, costly, and laborious annotation process conducted by a specialist.

\noindent \textbf{Breast Image Analysis Tasks}
Most AI applications in breast image analysis focus on the task of mass detection (annotating bounding boxes over regions of interest) and segmentation. Mass detection is often studied because of its direct applicability in Computer Aided Diagnosis (CAD) style software, because of the improved diagnostic capabilities provided by such models. Mass detection and segmentation in DM and DBT images is especially valuable for reducing clinical workloads~\citep{ZHANG202311,raya2021ai}. Mass detection and segmentation tasks often occur together, since segmentation models may contain a separate bounding box detection step, where any predicted regions of interest (ROIs) are classified by lesion presence/properties and then further segmented by the model, as in a Mask R-CNN~\citep{maskrcnn}. Alternatively, a model such as a U-Net can segment findings directly~\citep{ronneberger2015u}. Regardless of the model used, AI techniques have been applied to the task of segmentation and detection in both DM~\cite{seg_dm} and DBT modalities~\citep{BAI2021102049}, albeit with varying success, occasionally being prone to false-positive predictions or over-fitting~\citep{ZHANG202311}. AI analysis of non-mass breast tissues, such as fibroglandular and adipose tissues, remains under-explored, despite the utility of these tissues to identify potentially cancerous regions~\citep{ZHU2017e11, DL_density_segmentation}, leading to more precise diagnostics~\citep{dens}. To the best of our knowledge, there exist only a few works in segmenting non-lesion breast tissues~\citep{ver_1}, and among them, only two use deep learning methods~\citep{DL_density_segmentation, ver_2}. This scarcity is due, at least in part, to the severe dearth of annotations for these tissues in publicly available datasets (see Appendix for a summary of existing datasets).

\noindent \textbf{Synthetic Data} \label{sec:related_work_gen_modelling}
A potential way to mitigate data scarcity and class imbalance is to utilize synthetic data generation. Employing generative AI to produce these data, however, poses several shortcomings. First, the synthesized ground truth information is limited by what is captured in the observed data~\citep{badano2023stochastic}. Limitations stem from both the imaging system (for example, tissue orientations that cannot be differentiated because they lie along the system's null space) and human annotators (imperfect segmentation masks or bounding boxes). Secondly, generative models can perpetuate biases that exist within the data, marginalizing underrepresented demographics and fringe cases~\citep{zack2024assessing}. Ironically, synthetic data are most needed in instances where these biases exist. Finally, even when data restrictions are ignored, the finite expressivity and imperfect training process of the model itself results in considerable errors. For example, \cite{kelkar_2023} found that generative adversarial networks trained on simulated medical images failed to match their training sets in several key clinically relevant features. A recent work by \cite{Shumailov2024} also demonstrated that generative models trained over generations of synthetic data compounded errors and led to `model collapse'.

Our dataset relies on knowledge-based (KB) models which inherently incorporate domain expertise (``knowledge'' of physics and biology), which are used to procedurally create realistic images and annotations. Although KB models require rigorous domain-specific knowledge and intensive, realistic simulation, the pre-image ground-truth characteristics of a sample (for example, the underlying breast phantom) are precisely known. This allows for subgroup analyses, controlled comparison of different scan procedures, and control of dataset balance. Here, for example, we demonstrate how our KB data can analyze detection performance among different mass sizes and densities, information that is not available from patient data.

\section{Summary}
In this project, we demonstrate how knowledge-based (KB) models can be used as a source of synthetic data containing bounding box and pixel-level annotations for breast tissues in DM and DBT.  We publicly release this dataset to accelerate AI development in medical imaging applications. The model we use, developed by \cite{graff2016new}, consists of multiple components (glandular and adipose tissue, skin, ligaments, ducts, veins and arteries), and allows the insertion of lesions (see Fig.~\ref{fig:summary_figure}). By manipulating the properties of each component during the virtual rendering process, we can generate a variety of richly annotated examples. This ability is unique to the data generation models that decouple multi-scale components of the patient anatomy and sensor, unlike generative approaches that learn from imaging data alone~\cite{badano2023stochastic} (see Sec.~\ref{sec:related_work_gen_modelling} for more discussion). Our contributions are:

\begin{itemize}[leftmargin=*,itemsep=0pt,topsep=0pt]

\item We release T-SYNTH, a public synthetic dataset of paired DM (2D imaging) and DBT (3D imaging) images derived from a KB model, with pixel-level segmentation and bounding boxes of a variety of breast tissues.
\item We demonstrate how T-SYNTH can be used for subgroup analysis. Specifically, Faster-RCNN~\cite{shaoqing2015fasterrcnn} is trained for and evaluated for lesion detection in a balanced dataset; results reveal expected trends in subgroup performance in both DM and (C-View) DBT (e.g., less dense lesions are harder to detect). 
\item We train detection models on limited patient data in both DM and DBT (C-View), and show that augmenting training data with T-SYNTH can improve performance.
\end{itemize}

\section{Resource Availability}
\subsection{Data/Code Location}
All data and pretrained models are available on Hugging Face~\footnote{\url{https://huggingface.co/datasets/didsr/tsynth}}. We also uploaded all pretrained models along with their descriptions. Using these models, it is possible to post-process and reproduce all images used in this paper. The code used to train the models available on GitHub in T-SYNTH repository\footnote{\url{https://github.com/DIDSR/tsynth-release}}.

\subsection{Potential Use Cases}
The published data can be used to train detection or segmentation models in combination with patient data. As shown in the manuscript, when patient data is limited, the training can benefit from inclusion of synthetic data, especially for underrepresented subgroups.

\subsection{Licensing}
All data and code are released under the CC0-1.0 license. This permits unrestricted use, distribution, and reproduction in any medium, provided the original work is properly cited. Pretrained models are also shared under the same license.

\subsection{Ethical Considerations}
This work is using synthetically generated data and does not include any patient-identifiable information. Publication of synthetic data aims to promote transparency, reproducibility, and fairness in medical AI research.

\section{Methods}
The open-source Virtual Imaging Clinical Trials for Regulatory
Evaluation (VICTRE) pipeline~\citep{Badano2018victre} is used to simulate DM and DBT images with variations to breast density and mass properties, similar to the M-SYNTH dataset~\citep{sizikova2022improving}. The resulting datasets can be used for downstream AI analysis as discussed in Sec.~\ref{sec:downstream_analysis}. 

\subsection{Phantom Generation}
Controlled settings and their factors for phantom generation include: breast density: fatty, scattered, heterogeneous, and dense; lesion density: 1.0, 1.06, and 1.1 times the density of glandular tissue and lesion diameter: 5, 7, and 9 mm. 150 breast phantoms with lesions were generated for each factorial combination except for 9 mm lesions in heterogeneous and dense breasts, due to the smaller size of these denser breasts. Corresponding negative (lesion absent) samples were also generated for each positive sample, for a total of 9,000 images (4,500 lesion present and 4,500 lesion absent).
\paragraph{Lesion Growth and Insertion} \cite{sengupta2024situ} introduced a computational model for lesion growth of breast cancer within a three-dimensional voxelized breast model, accounting for stiffness of surrounding structures and considering both avascular and vascular phases of tumor development. The growth of lesions simulated with VICTRE pipeline. During the avascular phase, the tumor growth is limited by nutrient diffusion from surrounding tissues, leading to high interstitial pressure at the core and eventual necrosis due to oxygen deprivation. This triggers the release of tumor angiogenesis factors, promoting new blood vessel formation and transitioning the tumor into the vascular phase.

The growth dynamics are modeled using advection-reaction-diffusion equations. The model calculates interstitial tumor pressure and incorporates oxygen as a primary nutrient, influencing cell metabolism, angiogenesis, and metastasis. The tumor cells' proliferation is guided by local tissue stiffness, with softer tissues offering less resistance, thereby promoting tumor expansion in these regions. For imaging, the generated lesion models are incorporated into fatty breast models using voxel replacement techniques. The approach to modeling breast cancer lesion growth in silico, enabling the simulation of realistic and biologically-relevant lesion morphologies. Due to the model's ability to account for the mechanical properties of surrounding tissues, we used it in our research.

\subsection{Acquisition System Modeling}
\noindent \textbf{DBT Imaging} DBT images are created via a replica of the Siemens detector based on MC-GPU~\citep{badal2009accelerating}. DM images are produced using the GPU-accelerated Monte Carlo x-ray transport simulation code, replicating the Siemens Mammomat Inspiration system (matching those from the M-SYNTH work). 

\noindent \textbf{C-View Images} Majority of existing patient datasets are limited to C-View images (i.e., 2D images obtained from DBT projections) as they are less computationally intensive to analyze. We therefore perform comparative evaluation of T-SYNTH using lesion detection in C-View. Commercial algorithms for synthesizing C-View images from DBT data are proprietary. As such, we used the C-View approximation from \cite{klein2023synthesized}, which applies a spherical sharpening filter to the 3D volume and then taking the pointwise maximum along the compressed axis. For all instances, lesion bounding boxes were computed to encapsulate positive pixels in the known lesion mask.

\section{Experimental Setup}
We investigate T-SYNTH as tool for subgroup analysis in synthetic data (Sec.~\ref{sec:synth_subgroup_analysis}), augmenting limited patient datasets and investigations into how the composition of data affect downstream performance (Sec.~\ref{sec:impact_of_data_composition}). The Torchvision implementation of Faster R-CNN was used as the detection model, with a ResNet50 feature pyramid network pretrained on COCO~\citep{shaoqing2015fasterrcnn}. Five model instances each for DBT C-View and DM were trained and evaluated. Training loss was the summation of the classifier/objectness and box regression losses from the region proposal network and R-CNN. AdamW optimizer with a learning rate of 1e-4 was used, with a batch size of 24 images.  Models were tuned by measuring sensitivity on the validation set at a decision threshold of 0.5, which was checked every 100 training steps. Single V100s were used to train models through 3000 training steps for experiments with subsets of data and on 5000 training steps when we trained on all  data, which took about 5-8 hours to complete. We follow the conventions set in M-SYNTH and analyze models trained on synthetic data and patient data, as outlined below. Experiments are conducted for both DM and C-View DBT.

\subsection{Datasets} \label{sec:downstream_analysis}
\noindent \textbf{Synthetic Subgroup Analysis}
The subgroup factors we analyze for detection performance are breast density (4 levels), lesion density (3 levels) and lesion size (3 levels), using 300 distinct digital models (phantoms) for each possible subgroup, see Tab. 1. Note that unlike patient data which is unbalanced in positive and negative samples, T-SYNTH is balanced.

\noindent \textbf{Patient Data Augmentation} 
To explore data augmentation using T-SYNTH, five detection models were trained using only patient data, and five were trained with patient + synthetic data. EMBED was used as the patient (real) dataset~\cite{jeong2023emory}. Unfortunately, very few DBT (all C-View) images of fatty and dense breasts with findings are present in EMBED data. The available images were split into a training, validation and test set, such that each breast density was approximately represented equally in the validation and test sets, and no patients were shared between sets (see Appendix). Although a greater number of DM images are available, we maintained the same split for DM experiments for consistency.

Aside from the data used, all other training parameters are identical to the synthetic only training. A batch size of 26 images was used; 10 of which were from T-SYNTH in the patient + synth trials. Both methods had access to the entire patient training set. Patient images were sampled such that half contained at least one finding in each batch.

\begin{table}
\centering
\resizebox{0.8\linewidth}{!}{%
\footnotesize
\begin{tabular}{lcccc}
\toprule
Breast Density & Train Set & Val Set  & Test Set & Total\\
\midrule
Fatty     & 900/900 & 225/225 & 225/225 & 1350/1350 \\
Scattered & 900/900 & 225/225 & 225/225 & 1350/1350 \\
Hetero    & 600/600 & 150/150 & 150/150 &  900/900  \\
Dense     & 600/600 & 150/150 & 150/150 &  900/900  \\
\bottomrule
\textbf{Total} & \textbf{3000/3000} & \textbf{750/750} & \textbf{750/750}  & \textbf{4500/4500}\\ 
\bottomrule
\end{tabular}
}
\label{tab:tsynth_datasplit}
\caption{Distribution of \textbf{image} samples (+/-) from the T-SYNTH dataset (DM and DBT).}
\end{table}

\subsection{Evaluation}
Each model is evaluated by calculating the number of true positives (successfully detected lesions), false positives (regions falsely declared as a lesion), and false negatives (missed lesions) that a model outputs over the test set at varying decision thresholds. Similar to the DBTex detection challenge for DBT~\citep{konz_competition_2023}, a true positive was scored if the distance between the prediction and a ground truth boxes' centers was less than either half of the diagonal length of the ground truth box or 100 pixels. Also similar to this challenge, model sensitivity is reported in relation to the average number of false positives per image as a free-response receiver operator characteristic (FROC) curve. FROC is often preferred to precision-recall in medical screening contexts, where most exams do not contain any true findings, and the number of false positives per image more directly indicates the clinical workload incurred by the system. Finally, for each experiment, we retrain each model five times with different random seeds, and report the mean, minimum and maximum values of sensitivity at each FPR rate across the five models, respectively. Finally, differences in performance are analyzed across breast density, as well as lesion size and density subgroups. 

\section{Results and Discussions}
\subsection{T-SYNTH Data Visualization} \label{sec:visual_comp}
\begin{figure}[htb]
    \centering
    \includegraphics[width=1.0\linewidth]{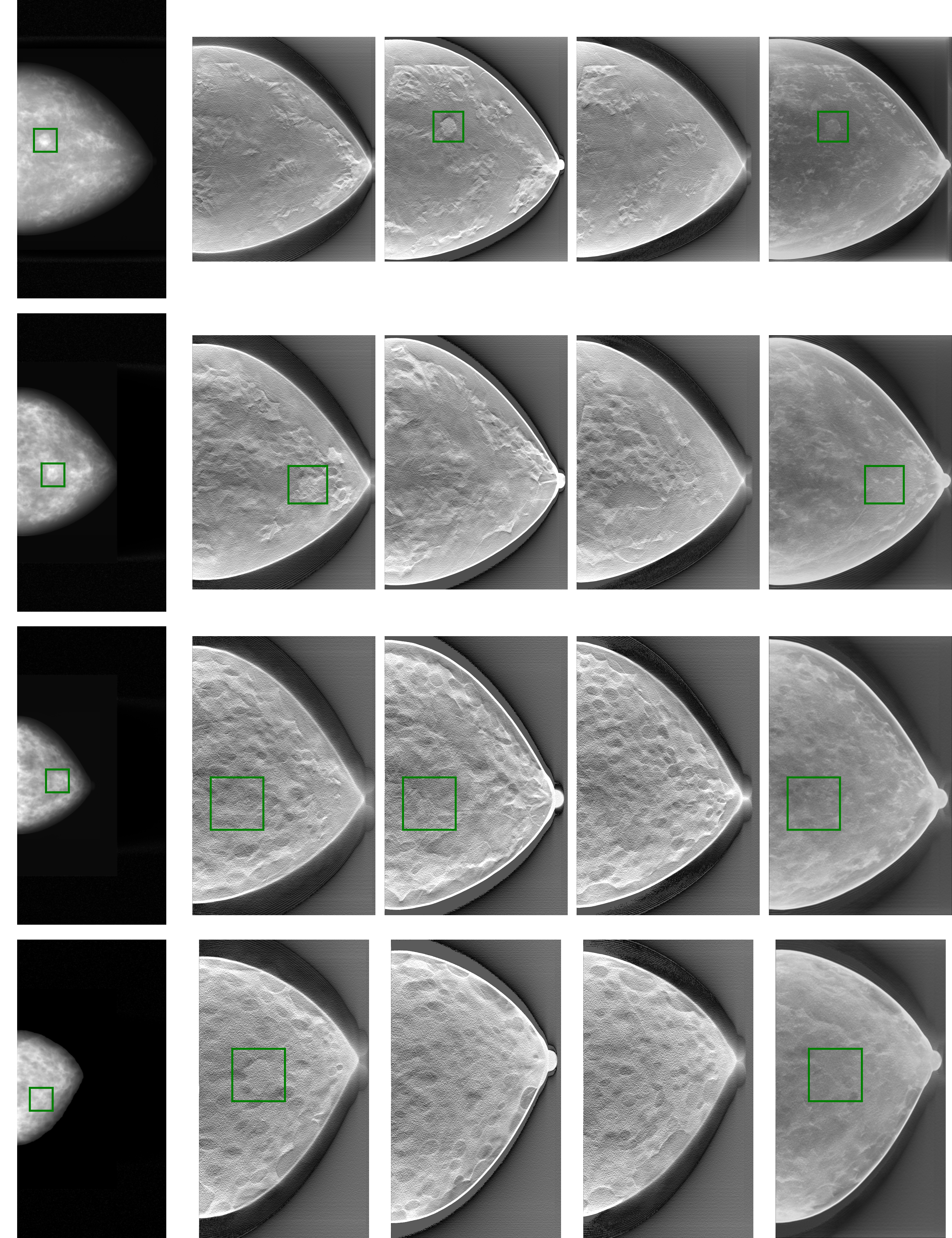}
    \caption{Example T-SYNTH images with varying breast densities (row 1: fatty, row 2: scattered, row 3: heterogeneous, row 4: dense). Column 1: DM, columns 2-4: DBT slices at quarter, half, and three-quarters of the total volume depth, Column 5: DBT C-View.}
    \label{fig:T-SYNTH_breast_density}
\end{figure}

Sample T-SYNTH images are shown in Fig.~\ref{fig:T-SYNTH_breast_density}. Visual differences can be observed between synthetic DM, DBT, and C-View images. These differences largely stem from the distinct reconstruction algorithms used, which notably differ from those employed with patient clinical data. Despite these variations, consistency in the underlying patient anatomy remains evident across image types.

Clinically relevant and expected trends related to lesion visibility can also be seen. Specifically, lesions are less distinct from the background fibroglandular tissue in higher breast density categories. This finding aligns with established medical literature, which notes that dense tissue regions can either obscure lesions, particularly projections with overlapping tissues, or mimic them. In the highest breast density considered (last row of Fig.~\ref{fig:T-SYNTH_breast_density}), DBT imaging provides a clearer visualization of lesions that were otherwise obscured by dense tissue in traditional DM images.

Similar visibility trends were observed across additional subgroup characteristics (see Appendix): smaller lesions consistently proved more difficult to detect, whereas smaller but denser lesions were relatively easier to identify. Consequently, T-SYNTH offers a robust paired synthetic DM and DBT dataset whose visual trends align closely with expectations derived from existing medical research. This consistency enhances our confidence in the dataset’s validity and its potential applicability for training and evaluation purposes in breast imaging research. 

\begin{figure}[htb]
\centering
  \begin{subfigure}{0.48\linewidth}
    \caption{EMBED Lesions}
    \includegraphics[width=0.9\linewidth]{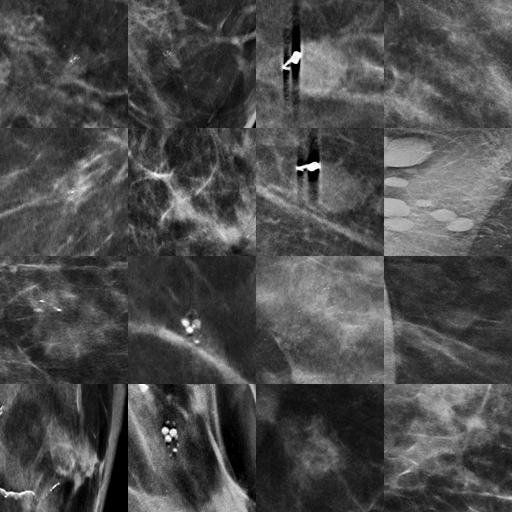}
  \end{subfigure}
  \begin{subfigure}{0.48\linewidth}
    \caption{T-SYNTH Lesions}
    \includegraphics[width=0.9\linewidth]{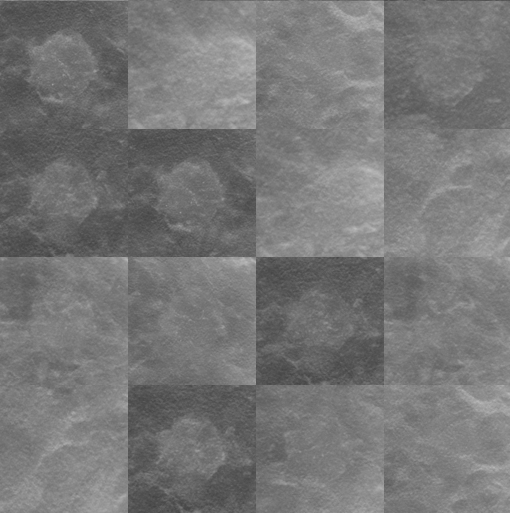}
  \end{subfigure}  
  \caption{Feature analysis of lesions in the EMBED and T-SYNTH datasets.}
  \label{fig:features}
\end{figure}

In Fig.~\ref{fig:features}, we qualitatively compare lesions in the EMBED and T-SYNTH datasets. We find that the EMBED bounding boxes may contain multiple lesions and may not be as consistently labeled as the bounding boxes in the T-SYNTH dataset.

\subsection{T-SYNTH Subgroup Analysis} \label{sec:synth_subgroup_analysis}
\begin{figure*}[htbp]
\centering
  \begin{subfigure}{0.30\linewidth}
    \includegraphics[width=0.9\linewidth]{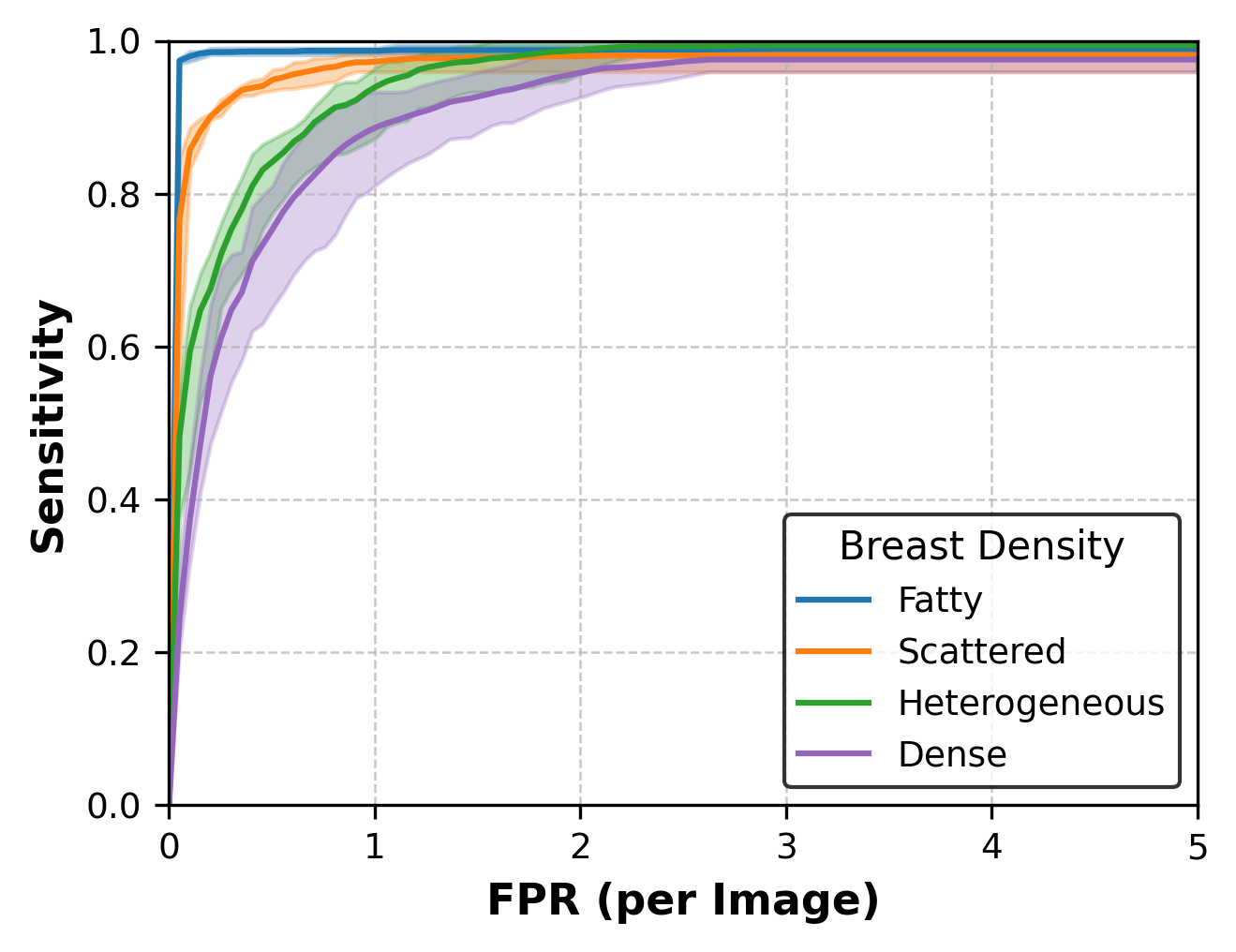}
    \caption{DBT - Breast Density}
  \end{subfigure}  
  \begin{subfigure}{0.30\linewidth}
    \includegraphics[width=0.9\linewidth]{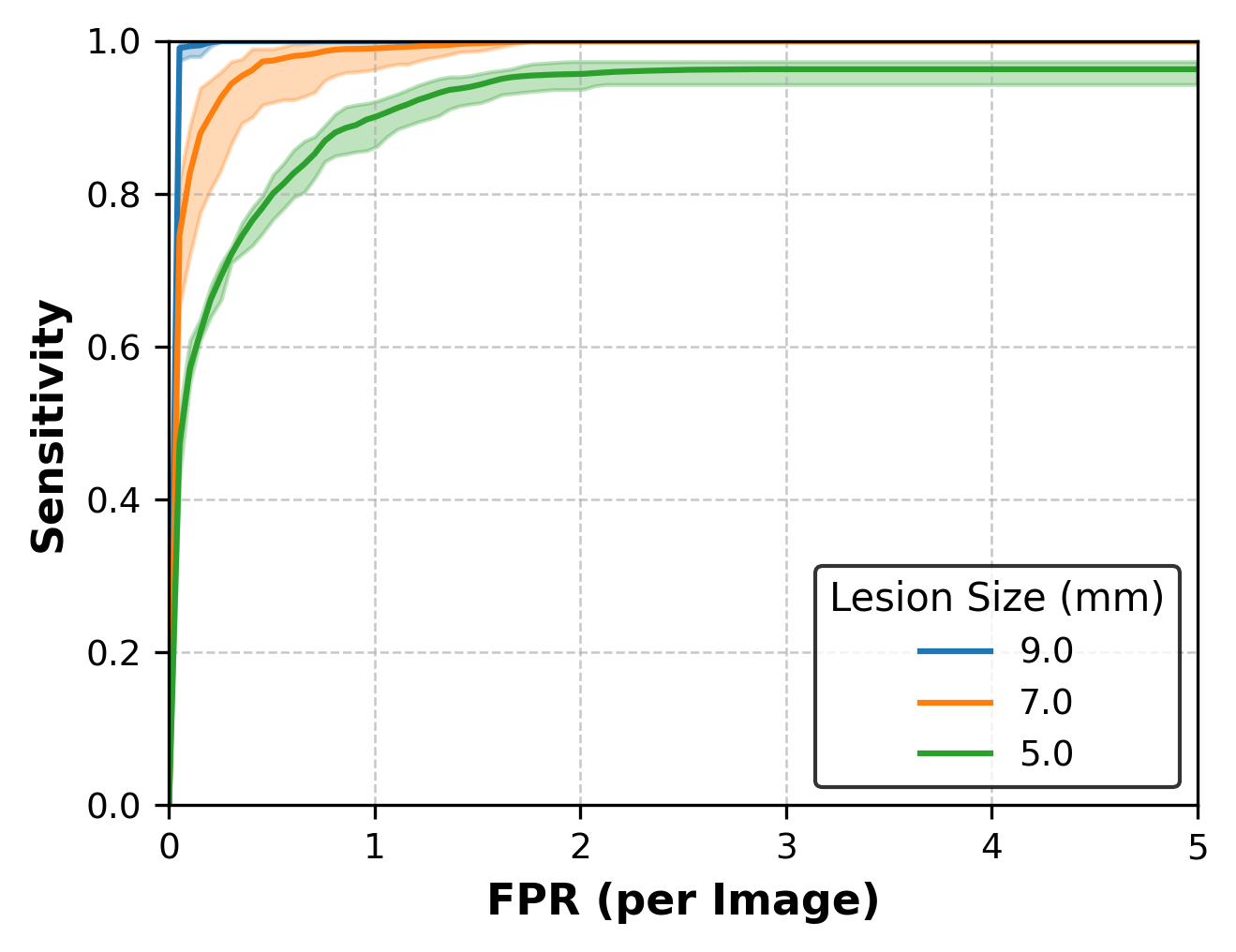}
    \caption{DBT -Lesion Size}
  \end{subfigure}
  \begin{subfigure}{0.30\linewidth}
    \includegraphics[width=0.9\linewidth]{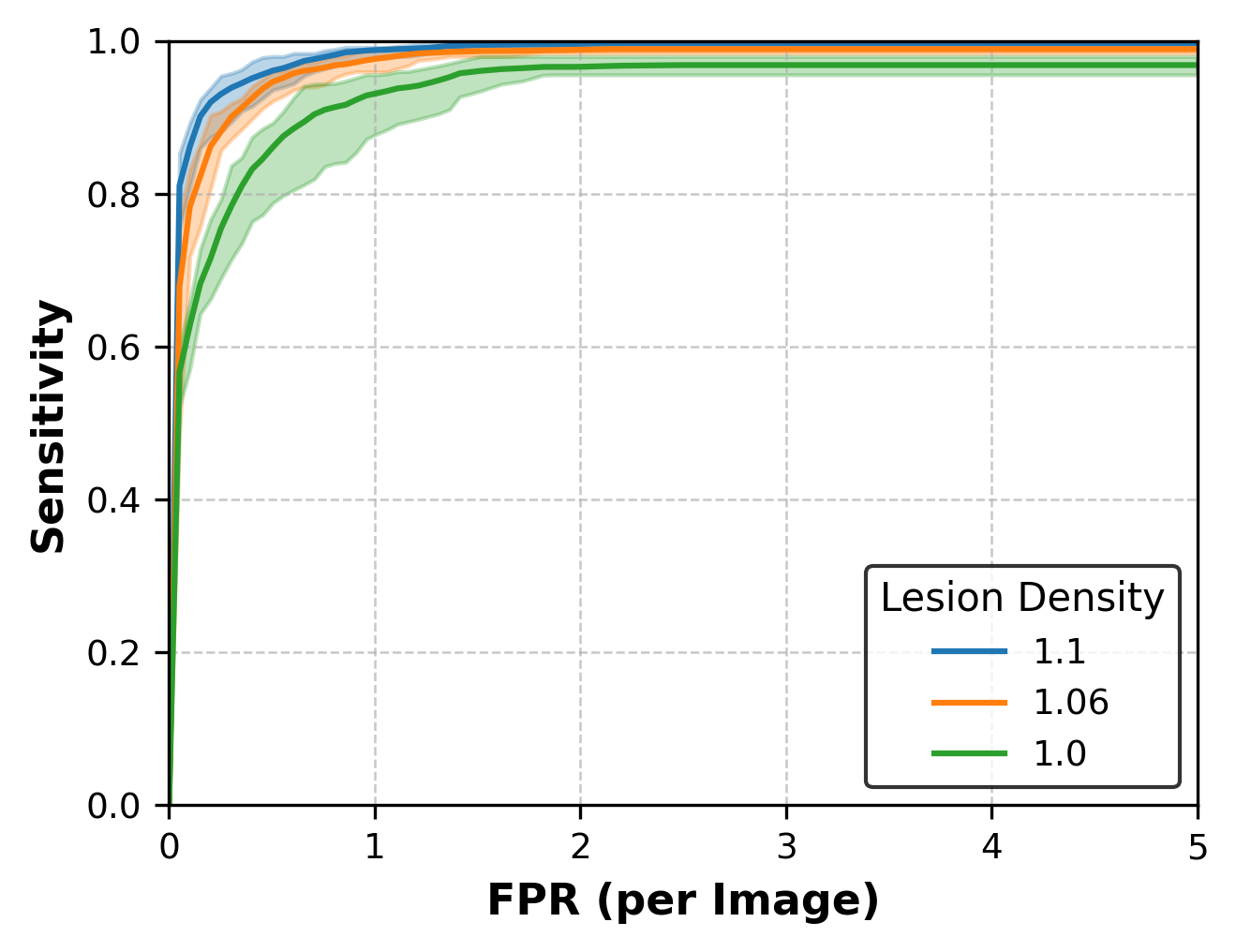}
    \caption{DBT -Lesion Density}
  \end{subfigure}

  \begin{subfigure}{0.30\linewidth}
    \includegraphics[width=0.9\linewidth]{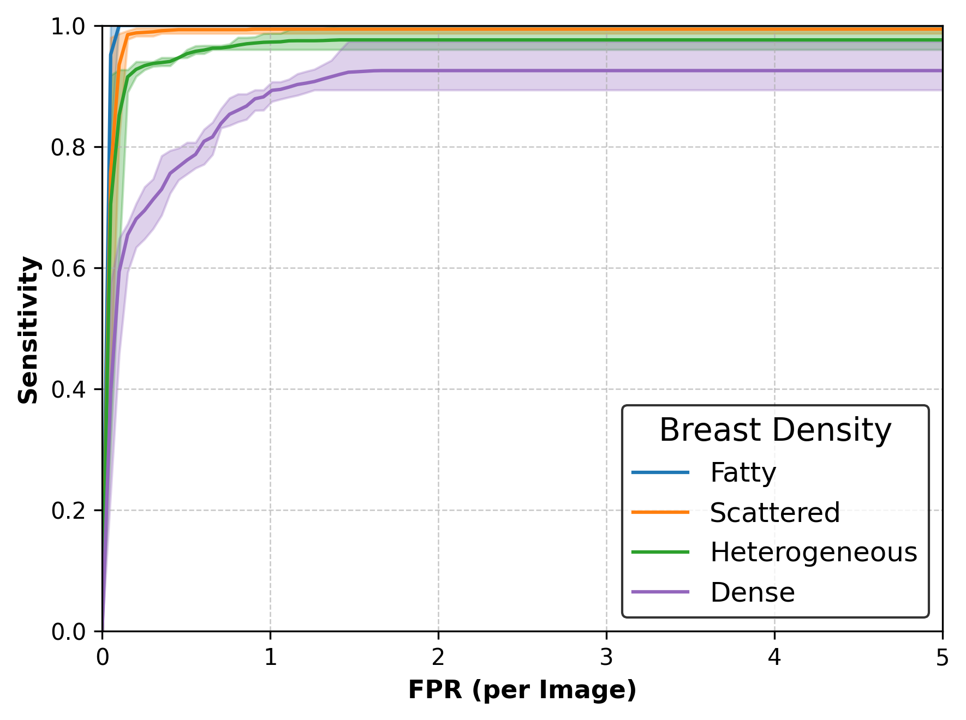}
    \caption{DM - Breast Density}
  \end{subfigure}  
  \begin{subfigure}{0.30\linewidth}
    \includegraphics[width=0.9\linewidth]{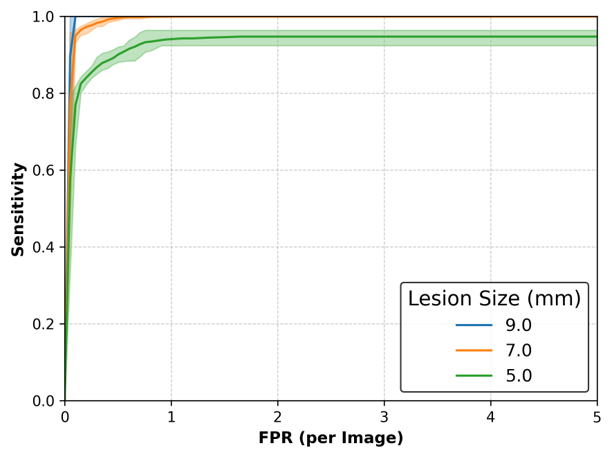}
    \caption{DM - Lesion Size}
  \end{subfigure}
  \begin{subfigure}{0.30\linewidth}
    \includegraphics[width=0.9\linewidth]{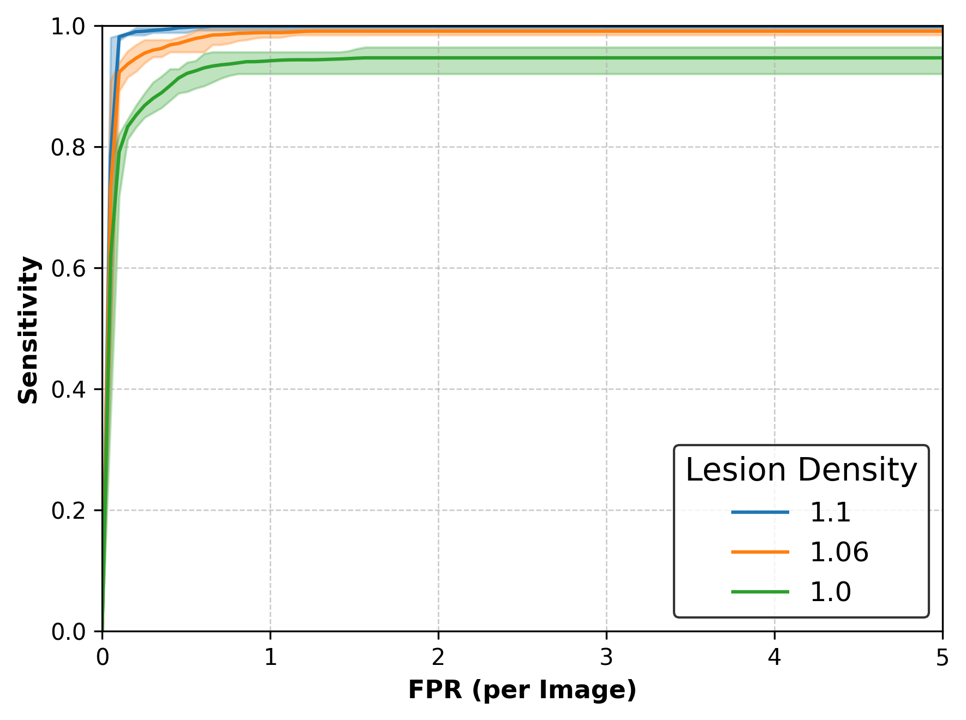}
    \caption{DM - Lesion Density}
  \end{subfigure}

  \caption{Synthetic Only Detection Results (Free response ROC Curve) comparing performance across subgroups: breast density, lesion size and density.}
  \label{fig:synthetic_only_froc}
\end{figure*}

In Fig.~\ref{fig:synthetic_only_froc}, we report detection results using FROC curves on models trained on synthetic data and evaluated on different subgroups in T-SYNTH. The x-axis represents the False Positive Rate (per image), and the Y-axis shows Sensitivity (True Positive Rate) defined as the proportion of correctly detected ground truth lesions. To generate these curves, each prediction was processed across a range of score thresholds (determined via the validation set) to compute detection outcomes. A detection was counted as a true positive if it matched a ground truth lesion and had not been matched previously by a higher-scoring prediction. Detections that did not match any ground truth lesions were counted as false positives. For each threshold, the total number of true positives was divided by the total number of ground truth lesions to obtain sensitivity (Se), while false positives (FP) were averaged per image. The shaded regions indicate the range (minimum to maximum) of sensitivity values across readers. In both DBT and DM examples, we find that performance improves with decreased density of fibroglandular tissue (i.e., breast density), increased lesion size, and increased lesion density, confirming the visual trends (see Sec.~\ref{sec:visual_comp}) and the overall clinical expectation. The spread of minimum values in DBT C-View images was slightly higher than in DM, likely due to additional noise. Because the analyzed subgroups are balanced and the nature of the phantoms control for other confounding factors, we can isolate these variables as the cause of differences in performance, boasting a potential advantage of T-SYNTH over patient data for subgroup analysis. 

\begin{figure*}[htbp]
\centering
  \begin{subfigure}{0.30\linewidth} 
    \includegraphics[width=0.9\linewidth]{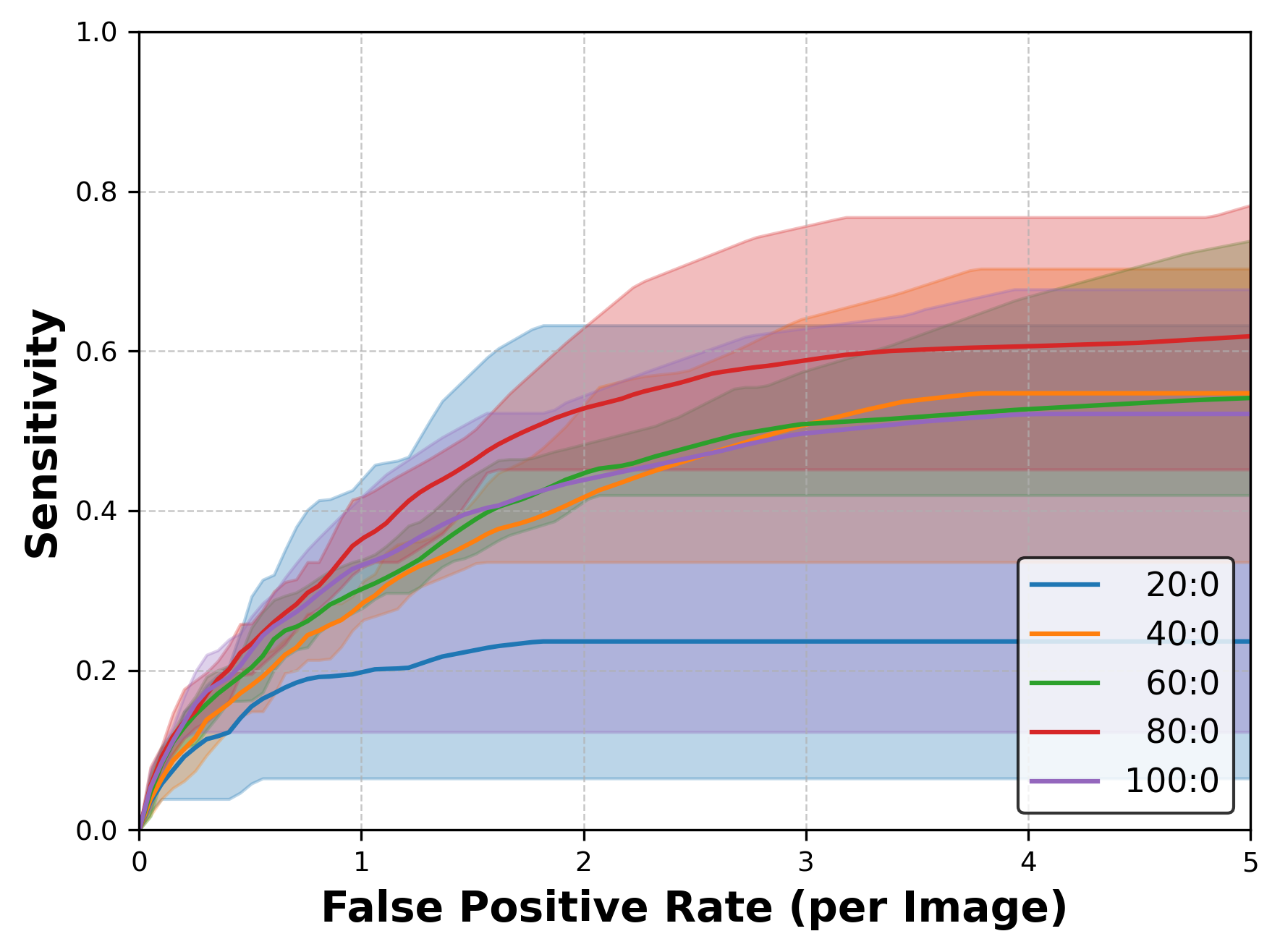}
    \caption{Patient Data Percent. Overall} \label{fig:sensitivity_froc_overall_real}
  \end{subfigure}  
  \begin{subfigure}{0.30\linewidth}
    \includegraphics[width=0.9\linewidth]{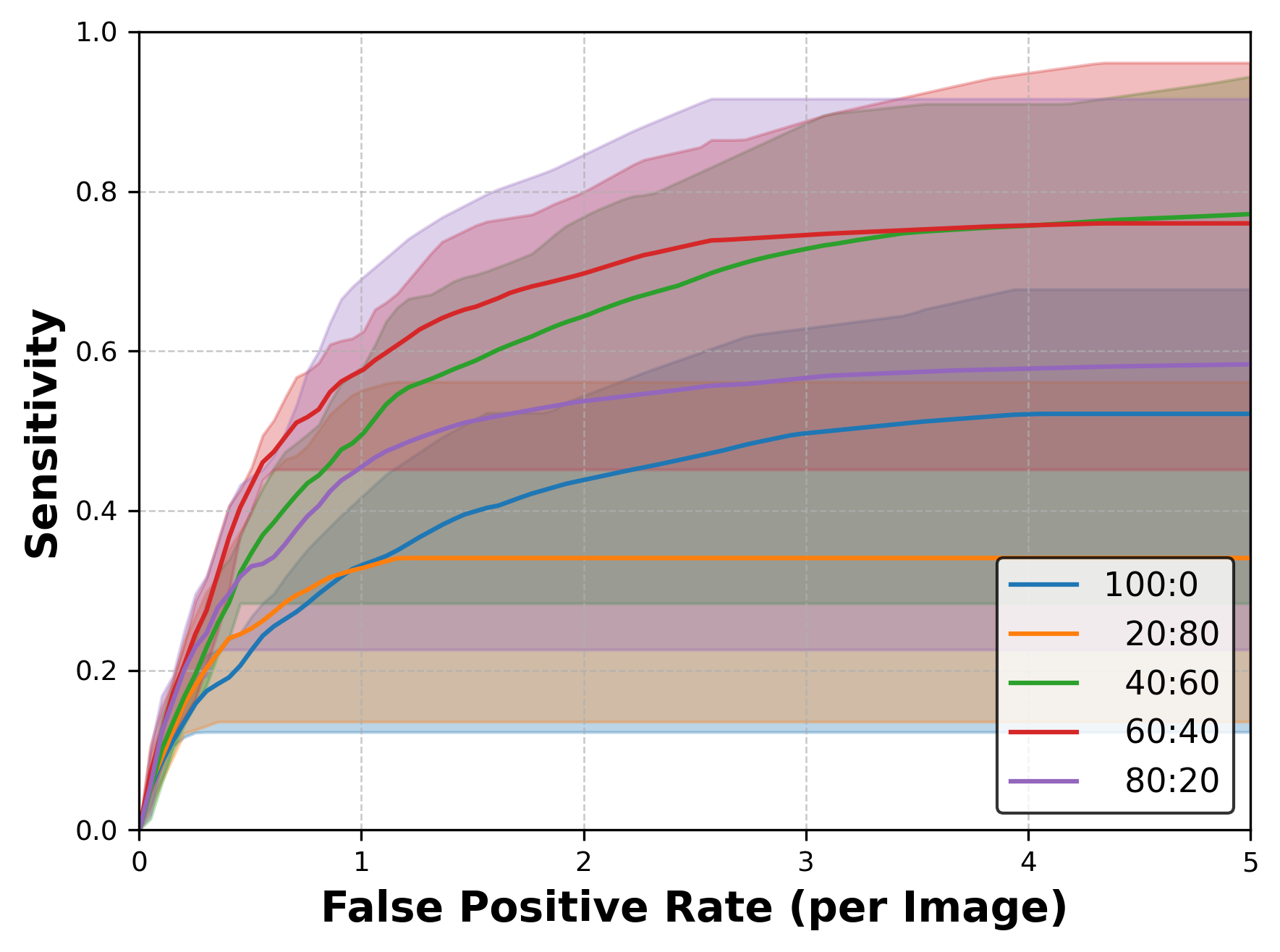}
    \caption{Replacement Overall} \label{fig:sensitivity_froc_overall_repl}
  \end{subfigure}
  \begin{subfigure}{0.30\linewidth}
    \includegraphics[width=0.9\linewidth]{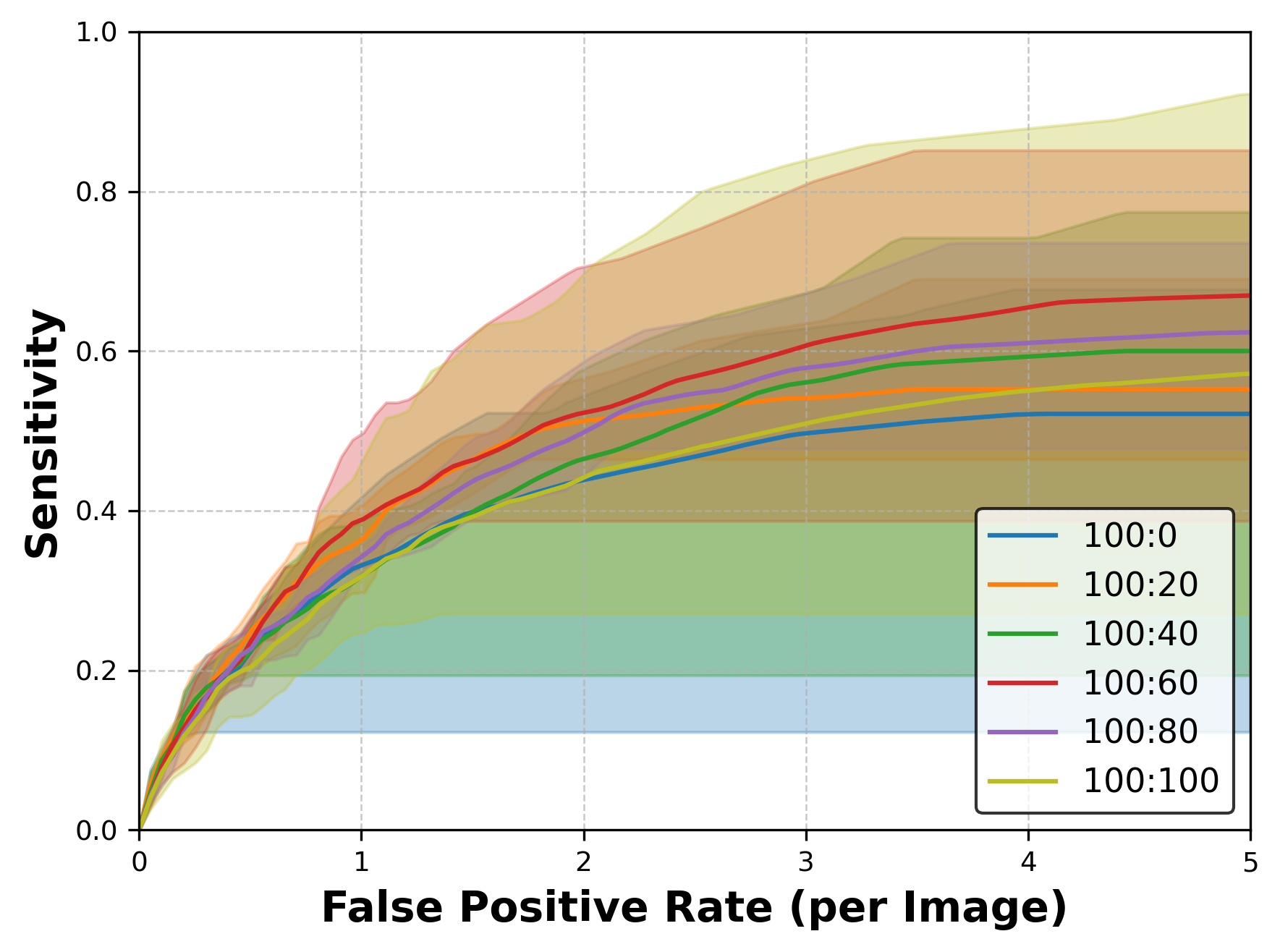}
    \caption{Addition Overall} \label{fig:sensitivity_froc_overall_add}
  \end{subfigure}

  \begin{subfigure}{0.30\linewidth} 
    \includegraphics[width=0.9\linewidth]{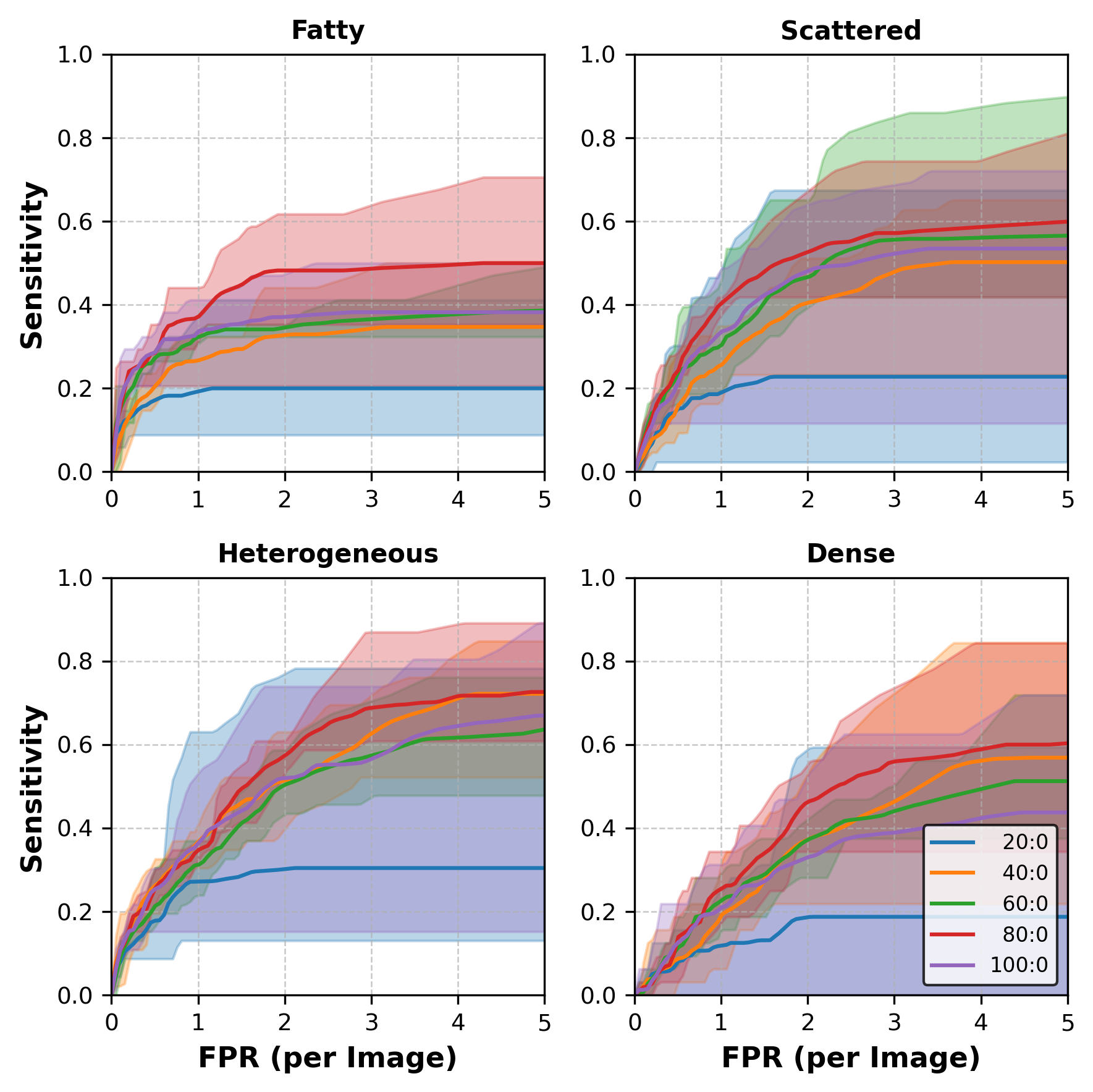}
    \caption{Patient Data Percent. Subgroup} \label{fig:sensitivity_froc_bd_real}
  \end{subfigure}  
  \begin{subfigure}{0.30\linewidth}
    \includegraphics[width=0.9\linewidth]{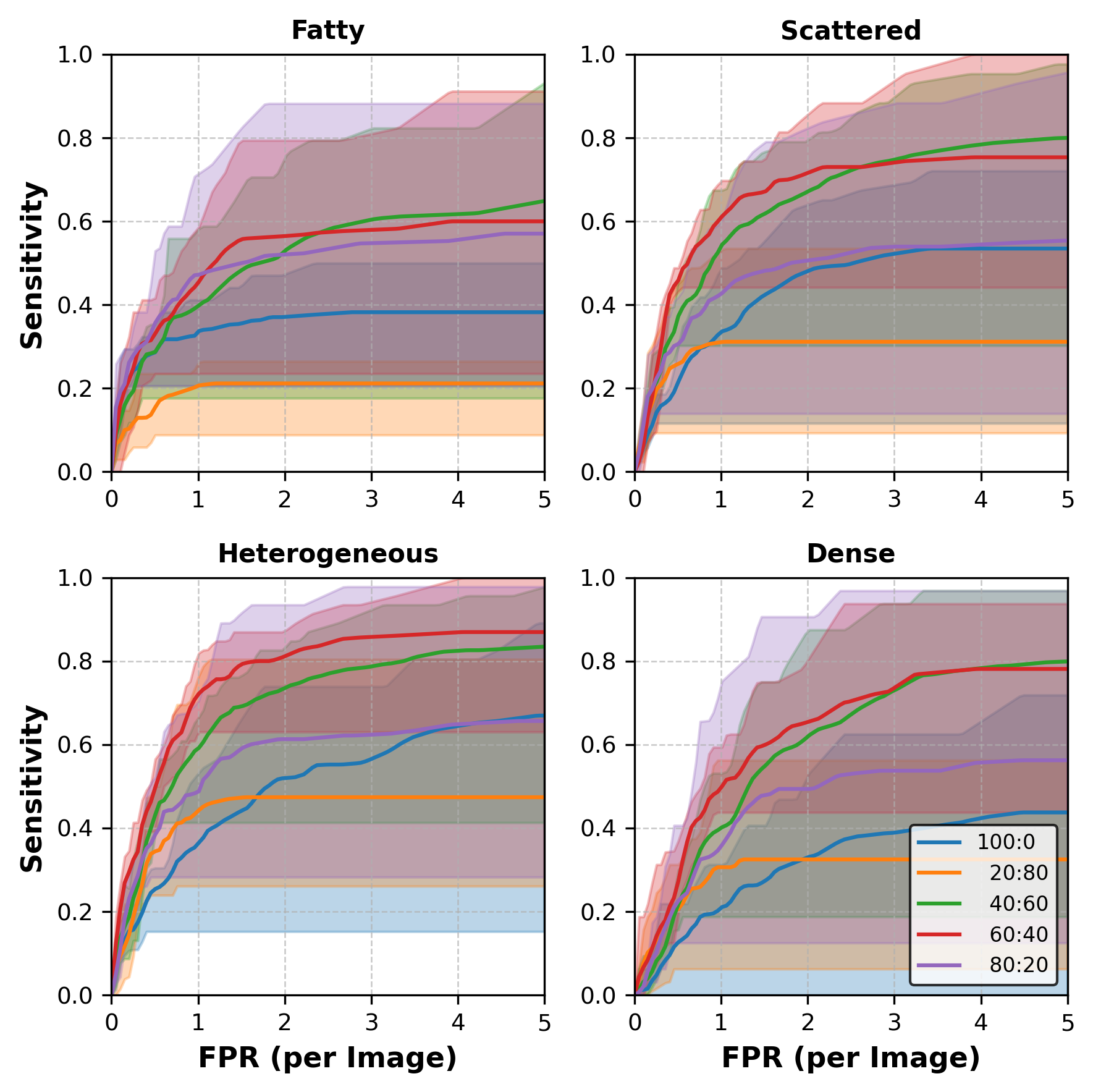}
    \caption{Replacement Subgroup} \label{fig:sensitivity_froc_bd_repl}
  \end{subfigure}
  \begin{subfigure}{0.30\linewidth}
    \includegraphics[width=0.9\linewidth]{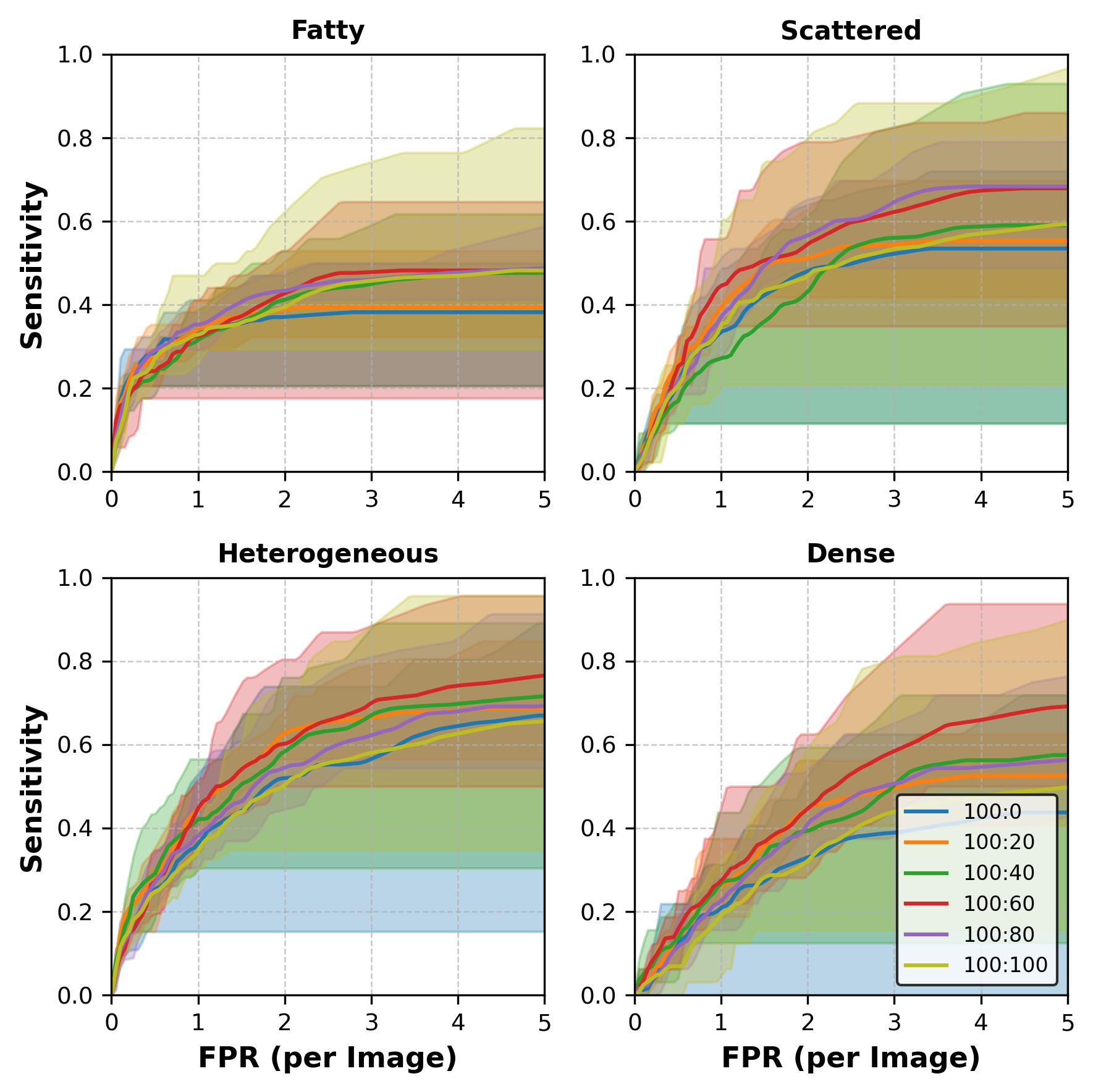}
    \caption{Addition Subgroup} \label{fig:sensitivity_froc_bd_add}
  \end{subfigure}

  \caption{\textit{Patient data percentage}: evaluation of when only a fraction of patient training data is available. \textit{Replacement}: evaluation of patient data with replacement of synthetic examples. \textit{Addition}: evaluation of patient data and addition synthetic examples}
  \label{fig:sensitivity_froc_controlled}
\end{figure*}

\subsection{Impact of Data Composition} \label{sec:impact_of_data_composition}
In this section, we evaluate the effect of data composition (patient:synthetic) examples on performance. In contrast to the above, we select a random subset of patient data to ensure that the number of patient and synthetic examples for training is fixed, but the composition of the training data changes. In both patient and synthetic cases, 100\% of data indicates the inclusion of 800 examples: 400 with lesions and 400 with no lesions. The test set here consists of 800 examples. As before, the FROC curves include five models, where solid line is average among readers and shaded area is shows the minimum and maximum, respectively. We run three types of experiments: 
\begin{itemize}[leftmargin=*,itemsep=0pt,topsep=0pt]
\item \textbf{Patient data percentage}: when model has only a fraction (20\%, 40\%, .., 100\%) of patient training data available (Fig.~\ref{fig:sensitivity_froc_overall_real} and Fig.~\ref{fig:sensitivity_froc_bd_real}).  
\item \textbf{Replacement}: when a proportion of patient data is replaced with synthetic data, e.g., patient (20\%) + synth (80\%) (Fig.~\ref{fig:sensitivity_froc_overall_repl} and Fig.~\ref{fig:sensitivity_froc_bd_repl}). 
\item \textbf{Addition}: when a proportion of synthetic data is added to the full available patient training set, e.g.,  patient (100\%) + synth (20\%)  (Fig.~\ref{fig:sensitivity_froc_overall_add} and Fig.~\ref{fig:sensitivity_froc_bd_add}).
\end{itemize}

From the patient data percentage experiment, we find that average performance drops significantly when only 20\% of patient data is available, but other experiments with different data percentage 40\%, 60\% and 80\% perform in a similar range. For fatty and scattered breasts, there is a drop from using only 40\% of data, but for dense and heterogeneous breasts, there is not. When a proportion of patient data is replaced with synthetic data and compared with the case when all patient data is present (100:0), we find that replacing 40\% to 80\% of patient data results in a marked improvement in performance over patient data only. Finally, we evaluate how FROC performance on patient test data (EMBED) is affected by the addition of synthetic examples (from T-SYNTH) and find that addition of T-SYNTH data improves performance at all levels of the image FPR, particularly for scattered and dense breast examples (see Appendix).

\subsection{Comparisons to a Generative AI Model} \label{sec:gen_ai_comparison}
We also compare the performance of T-SYNTH to stable diffusionSDXL~\citep{podell2023sdxl}, which generates images of 1024x1024 resolution. We generate images using this model with and without finetuning, where finetuning is done using the textual inversion~\citep{gal2022image} personalization method using a subset of 16 lesions from the patient dataset. We report results using the following controlled sets of training data: T-SYNTH, EMBED baseline, Subset EMBED baseline, Diffusion, and Finetuned Diffusion (see Appendix).

\begin{figure}[htbp]
\centering
  \begin{subfigure}{1.0\linewidth}
    \caption{Overall Performance}
    \includegraphics[width=0.9\linewidth]{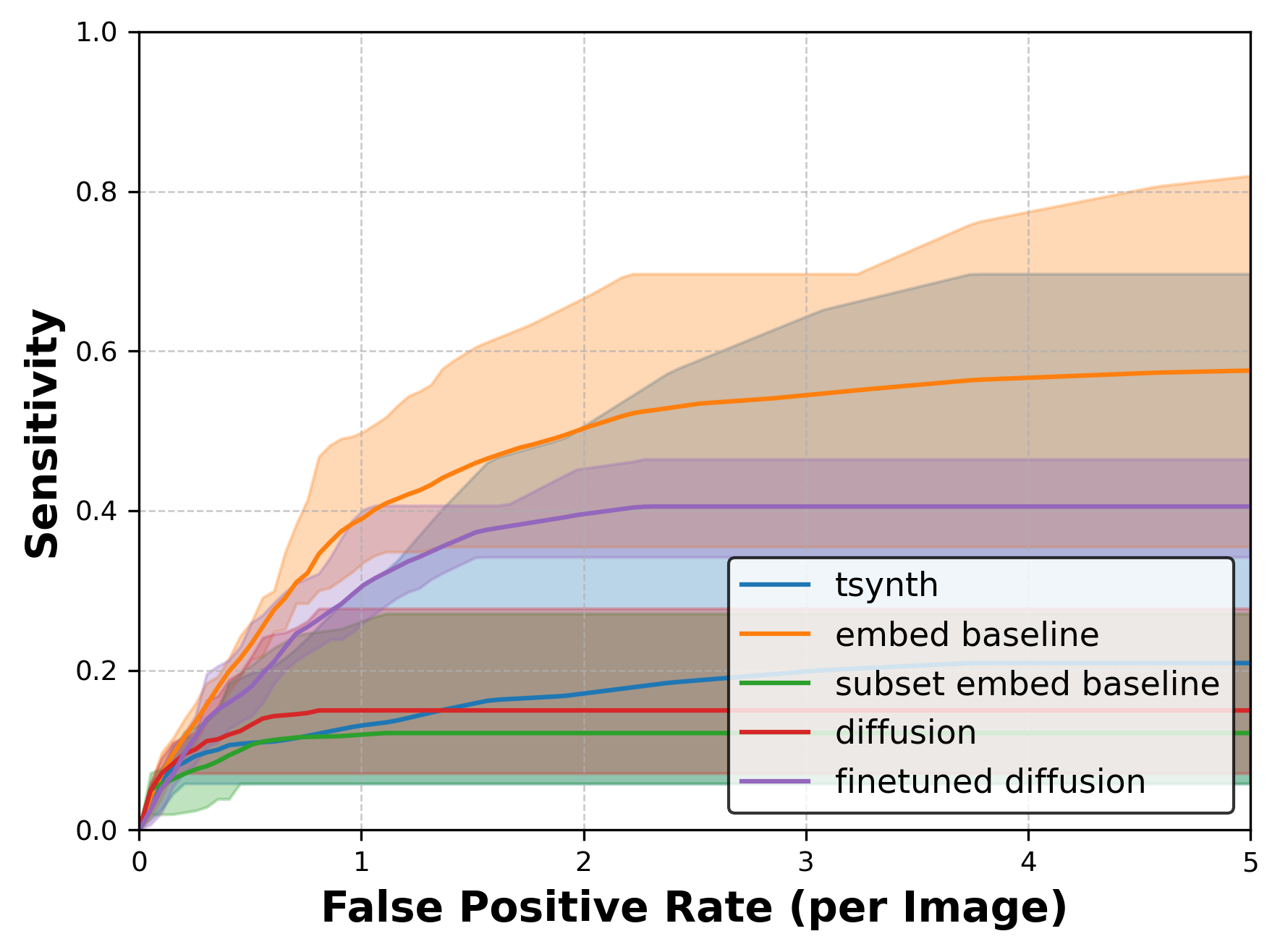}
    \label{fig:froc_diffusion_overall}
  \end{subfigure}  \\
  \begin{subfigure}{1.0\linewidth}
    \caption{Breast Density Subgroup Performance}
    \includegraphics[width=0.9\linewidth]{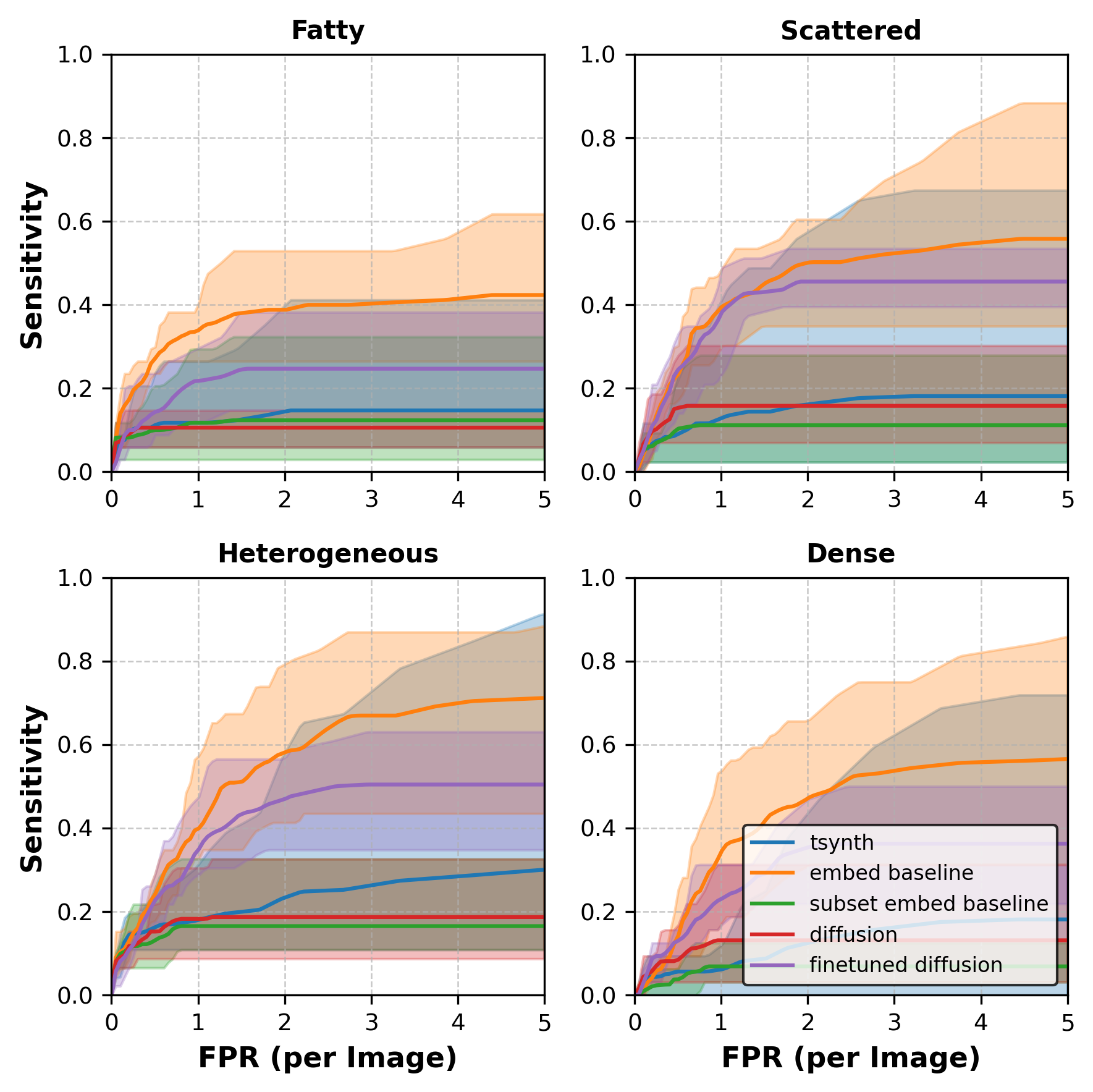}
    \label{fig:froc_diffusion_subgroup}
  \end{subfigure}
  \caption{Comparison of T-SYNTH, diffusion and finetuned diffusion synthetic data supplementation.}
  \label{fig:froc_diffusion}
\end{figure}

At a high level, the idea is to evaluate whether T-SYNTH and diffusion can supplement positive cases (with lesions present) which are underrepresented in the EMBED dataset. Our experiments indicate that both types of synthetic images (diffusion and T-SYNTH) improve performance over the under-sampled patient baseline. In the overall performance comparison (see Fig.~\ref{fig:froc_diffusion_overall}), the diffusion model without finetuning performs worse than T-SYNTH while the finetuned diffusion model performs better than T-SYNTH. Note that this is expected since T-SYNTH distributions have not been matched to patient samples in EMBED. Similar comparative performance trends are observed across subgroups (see Fig.~\ref{fig:froc_diffusion_subgroup}), although the magnitude of improvement is different across the four considered breast densities. 

\section{Limitations and Future Work}
A number of limitations exist with T-SYNTH as well as our analysis, which can be improved in future iterations. First, there is a domain gap between T-SYNTH and patient DM and DBT images. Much of this gap is due to differences in image reconstruction and postprocessing algorithms, which can greatly impact background intensity, picture sharpness, and tissue contrasts. It is important to mitigate these differences as best as possible, although this is challenging because commercial algorithms that are proprietary. Shifts can be addressed either by adjusting KB parameters, or domain shift mitigation techniques~\citep{billot2023synthseg, liu2022deep}. There is also room for improvement in the lesion model used in T-SYNTH, which does not currently capture. We followed a similar generation methodology as that of M-SYNTH dataset~\citep{sizikova2022improving} to enable a direct comparison between c-view and mammography, which can be further improved.

In addition, patient datasets, including EMBED, feature complex annotations of findings that are not limited to just masses. For example, classifications and focal asymmetries are noted, and these findings are often classified by their suspected severity via the BIRADs scale. Thus, our model can be expanded in future work to include other important abnormalities such as calcifications, as well as an improved lesion model that captures diversity of both expression and cancer severity that is seen in patient datasets. 

Finally, future work can also expand our analysis to the 3D DBT images in T-SYNTH. Currently, a robust public dataset of 3D DBT images for comparison, with rich annotations such as breast density classifications, is lacking (further highlighting the need for synthetic data). Additionally, while our preliminary findings noted improvement in detection performance on patient data when training data was augmented, there is substantial variation in results between test trials. This is likely due to the relatively small test and validation sets, which are limited by the relatively few number of fatty and dense breast samples containing findings.

\acks{We thank Kenny Cha, Mike Mikailov, Mohammad Akhonda for providing help with computational experiments' setup, and all members of the (DIDSR/OSEL/CDRH/FDA) team for helpful suggestions.}

\ethics{The work follows appropriate ethical standards in conducting research and writing the manuscript, following all applicable laws and regulations regarding treatment of animals or human subjects.}

\coi{The authors have no conflict of interests to declare.}

\clearpage

\FloatBarrier

\clearpage

\FloatBarrier

\clearpage
\section{Appendix}
\subsection{Additional T-SYNTH Visualizations}
\begin{figure}[htbp]
    \centering
    \includegraphics[width=1.0\linewidth]{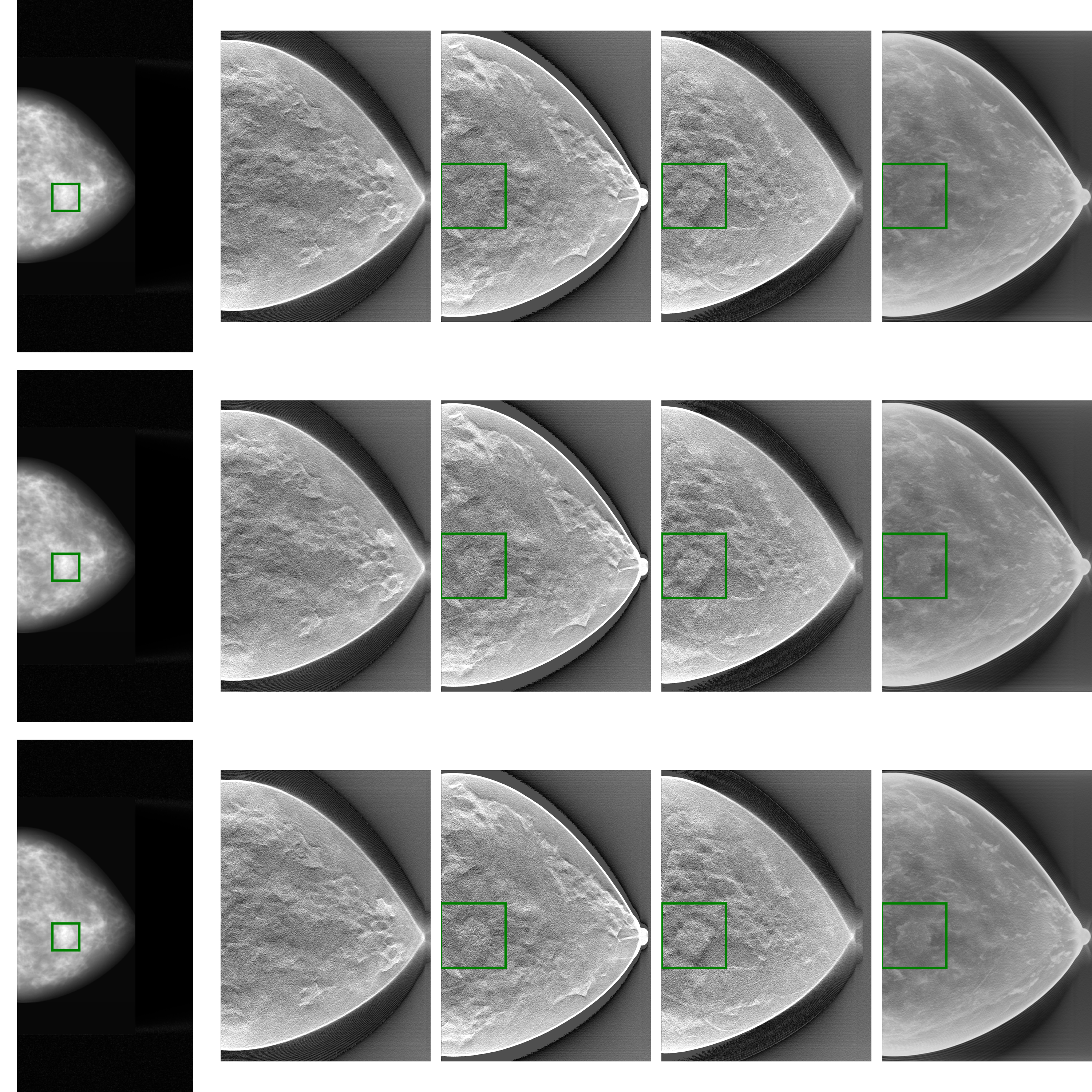}
    \caption{Example T-SYNTH images with varying lesion densities, as measured relative to fibroglandular density (row 1: 1.0, row 2: 1.06, row 3: 1.1). Column 1: DM, columns 2-4: DBT slices at quarter, half, and three-quarters of the total volume depth, column 5: DBT C-View.}
    \label{fig:TSYNTH_lesion_density}
\end{figure}

\begin{figure}[htbp]
    \centering
    \includegraphics[width=1.0\linewidth]{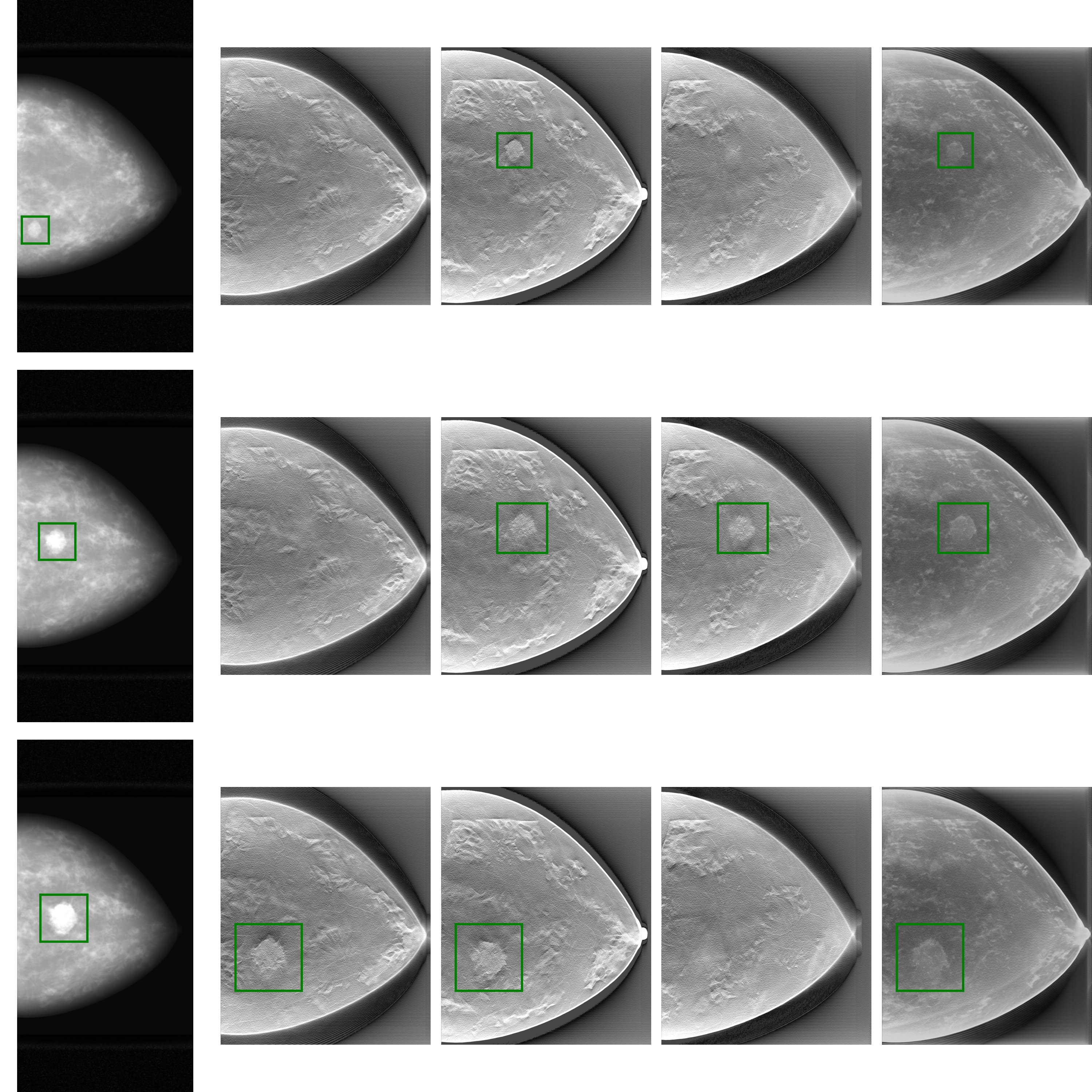}
    \caption{Example T-SYNTH images with varying lesion sizes (row 1: 5mm, row 2: 7mm, row 3: 9mm). Column 1: DM, columns 2-4: DBT slices at quarter, half, and three-quarters of the total volume depth, column 5: DBT C-View.}
    \label{fig:TSYNTH_lesion_size1}
\end{figure}

In Fig.~\ref{fig:TSYNTH_lesion_density}, we include examples of T-SYNTH images with varying lesion densities. In Fig.~\ref{fig:TSYNTH_lesion_size1}, we include examples of T-SYNTH images with varying lesion sizes. 

\subsection{Dataset Splits}
In Tab.~\ref{tab:embed_datasplit} and Tab.~\ref{tab:embed_datasplit_patient}, we provide a distribution of image and patient samples from the EMBED dataset. In Tab.~\ref{tab:tsynth_datasplit_patient}, we provide a distribution of patient samples from the T-SYNTH dataset (the corresponding image samples table can be found in the main paper). The difference across patient and images samples is that the same patient may contain multiple images (e.g., from different time points).

\begin{table}[htbp]
\resizebox{\linewidth}{!}{%
\centering
\footnotesize
\begin{tabular}{lcccc}
\toprule
Breast Density & Train Set & Val Set  & Test Set & Total\\
\midrule
Fatty     & 53/8614 & 34/34 & 29/29 & 116/8977\\
Scattered & 630/33812 & 35/35 & 38/38 & 703/33885 \\
Hetero    & 660/33391 & 34/34 & 38/38 & 732/33463 \\
Dense     & 33/4888 & 34/34 & 31/31 & 98/4953 \\
\bottomrule
\textbf{Total} & \textbf{1346/80705} & \textbf{137/137} & \textbf{136/136}  & \textbf{1649/81278}\\  
\bottomrule
\end{tabular}
}
\caption{Distribution of \textbf{image} samples (+/-) from the EMBED dataset (DM and DBT).}
\label{tab:embed_datasplit}
\end{table}

\begin{table}[htbp]
\resizebox{\linewidth}{!}{%
\centering
\footnotesize
\begin{tabular}{lcccc}
\toprule
Breast Density & Train Set & Val Set  & Test Set & Total\\
\midrule
Fatty     & - & - & - & - \\
Scattered & - & - & - & - \\
Hetero    & - & - & - & - \\
Dense     & - & - & - & - \\
\bottomrule
\textbf{Total} & \textbf{723/8334} & \textbf{73/0} & \textbf{78/0}  & \textbf{874/8334}\\  
\bottomrule
\end{tabular}
}
\caption{Distribution of \textbf{patient} samples (+/-) from the EMBED dataset (DM and DBT).}
\label{tab:embed_datasplit_patient}
\end{table}

\begin{table}[htbp]
\resizebox{\linewidth}{!}{%
\centering
\footnotesize
\begin{tabular}{lcccc}
\toprule
Breast Density & Train Set & Val Set  & Test Set & Total\\
\midrule
Fatty     & 100/100 & 25/25 & 25/25 & 150/150 \\
Scattered & 100/100 & 25/25 & 25/25 & 150/150 \\
Hetero    & 100/100 & 25/25 & 25/25 & 150/150 \\
Dense     & 100/100 & 25/25 & 25/25 & 150/150 \\
\bottomrule
\textbf{Total} & \textbf{400/400} & \textbf{100/100} & \textbf{100/100}  & \textbf{500/500}\\  
\bottomrule
\end{tabular}
}
\caption{Distribution of \textbf{patient} samples (+/-) from the T-SYNTH dataset (DM and DBT). Examples with same phantom but different lesion size or lesion density are treated as the same patient.}
\label{tab:tsynth_datasplit_patient}
\end{table}

\subsection{Additional FROC Analysis on Data Replacement}
\begin{figure}[htbp]
    \centering
    \includegraphics[width=1\linewidth]{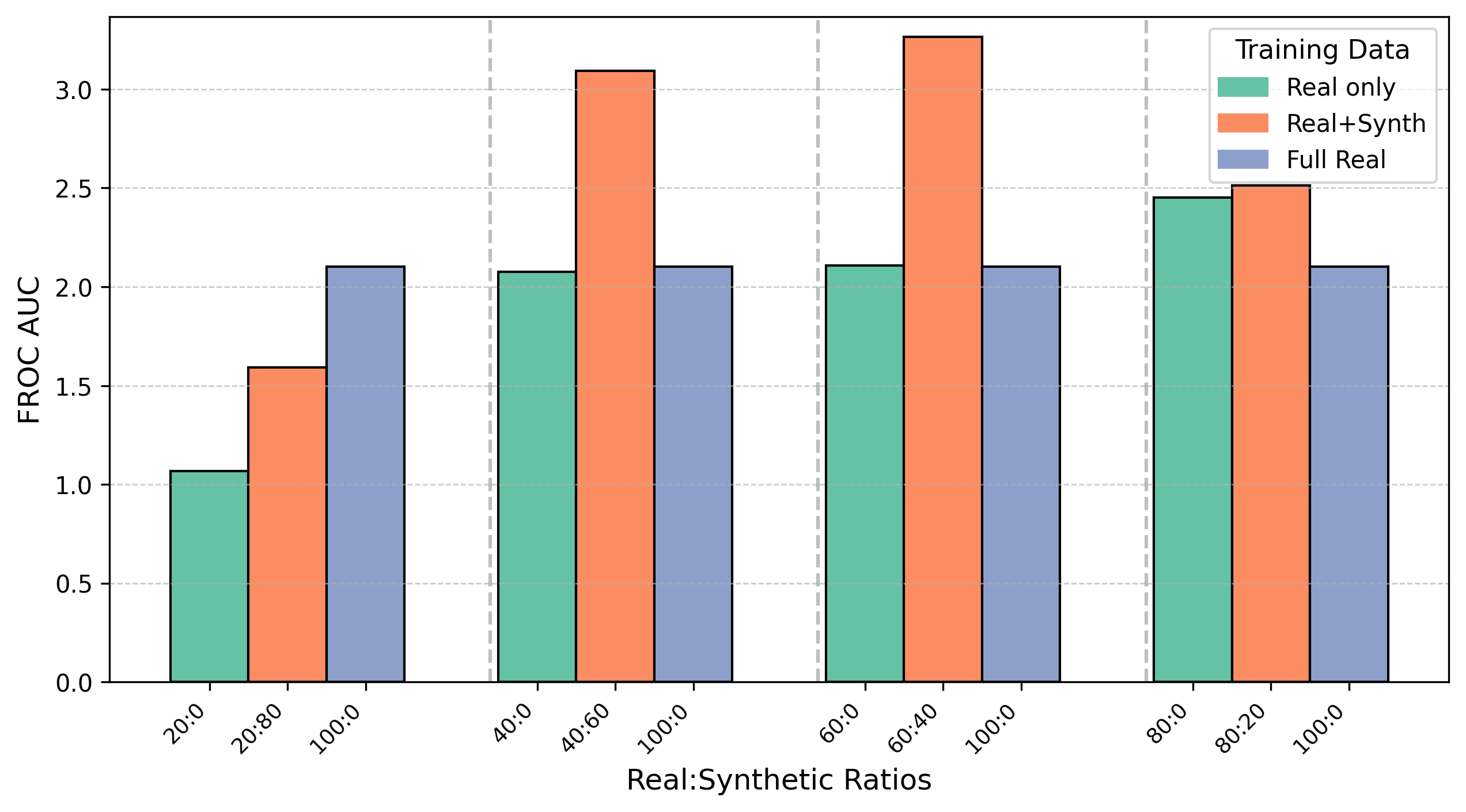}
    \caption{Comparison of FROC AUC scores across different real-to-synthetic training data ratios. Each bar represents the average performance across five runs. The \textcolor{blue}{blue} bar - 100\% patient data is included in each group for reference. The \textcolor{green}{green} bar is different proportion of patient data and \textcolor{orange}{orange} is replacement in different proportion with T-SYNTH data.}
    \label{fig:FROC_AUC}
\end{figure}

Finally, in  Fig.~\ref{fig:FROC_AUC} we evaluate the effect of different patient-to-synthetic training data ratios on detection performance, measured by the area under FROC (FROC-AUC). For each group, green bars represent models trained on partial patient training data relative to the full patient training dataset (800 images, consisting of 400 examples with lesions and 400 examples without lesions). Orange bars represent models trained on a mix of patient and synthetic training data which also has 800 images, but contains proportions (patient:synthetic) of patient and synthetic data. Blue bars show performance of models trained on the full patient dataset. We find that replacing nearly 60\% of patient data with synthetic examples achieves the same average level of performance as using the full patient training dataset.

\subsection{Comparison of DM and DBT}
\begin{figure*}[htbp]
\centering
  \begin{subfigure}{0.48\linewidth}
    \caption{DM Overall Result}
    \includegraphics[width=0.9\linewidth]{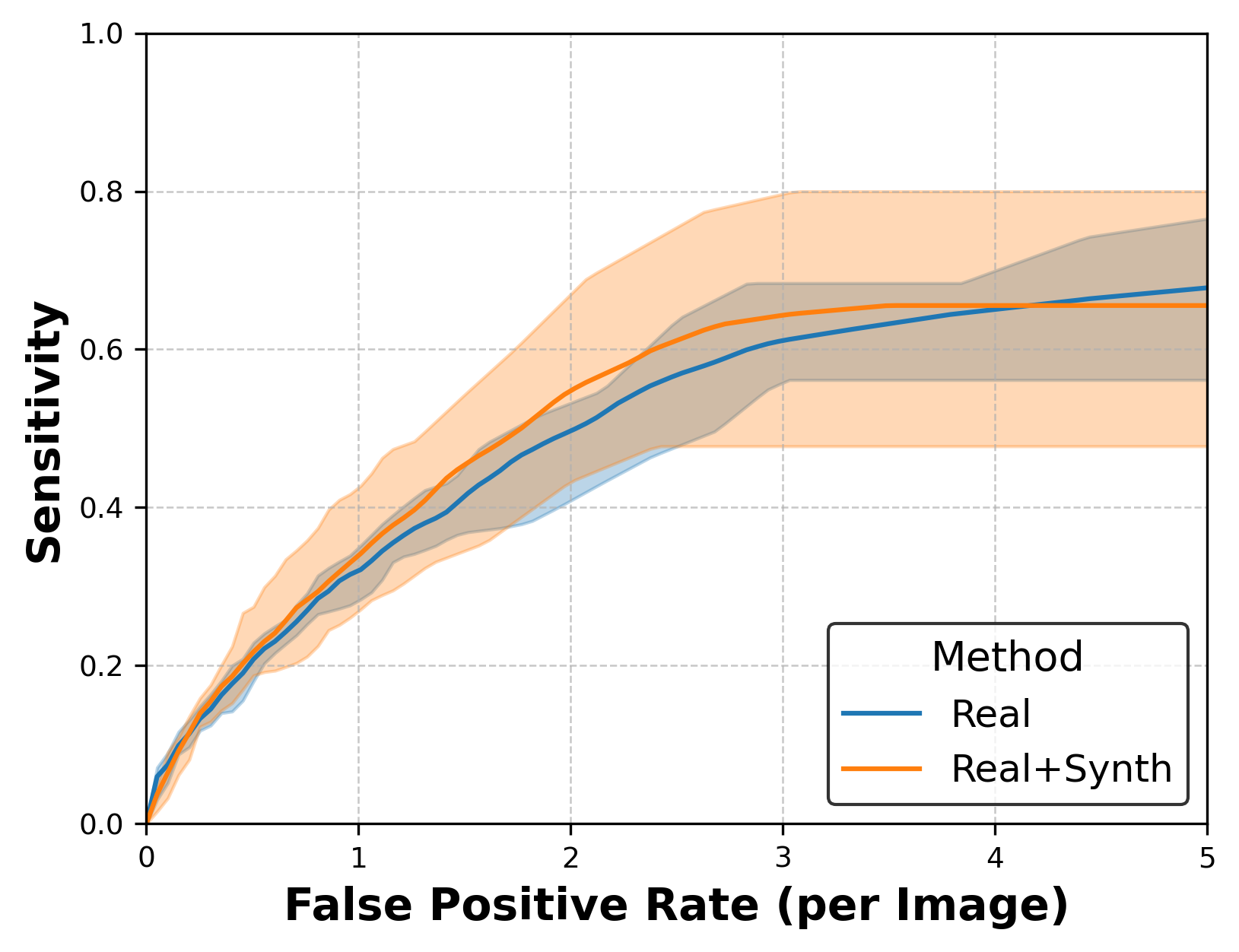}
    \label{fig:froc_real_synth_dm_overall}
  \end{subfigure}
  \begin{subfigure}{0.48\linewidth}
    \caption{DBT Overall Result}
    \includegraphics[width=0.9\linewidth]{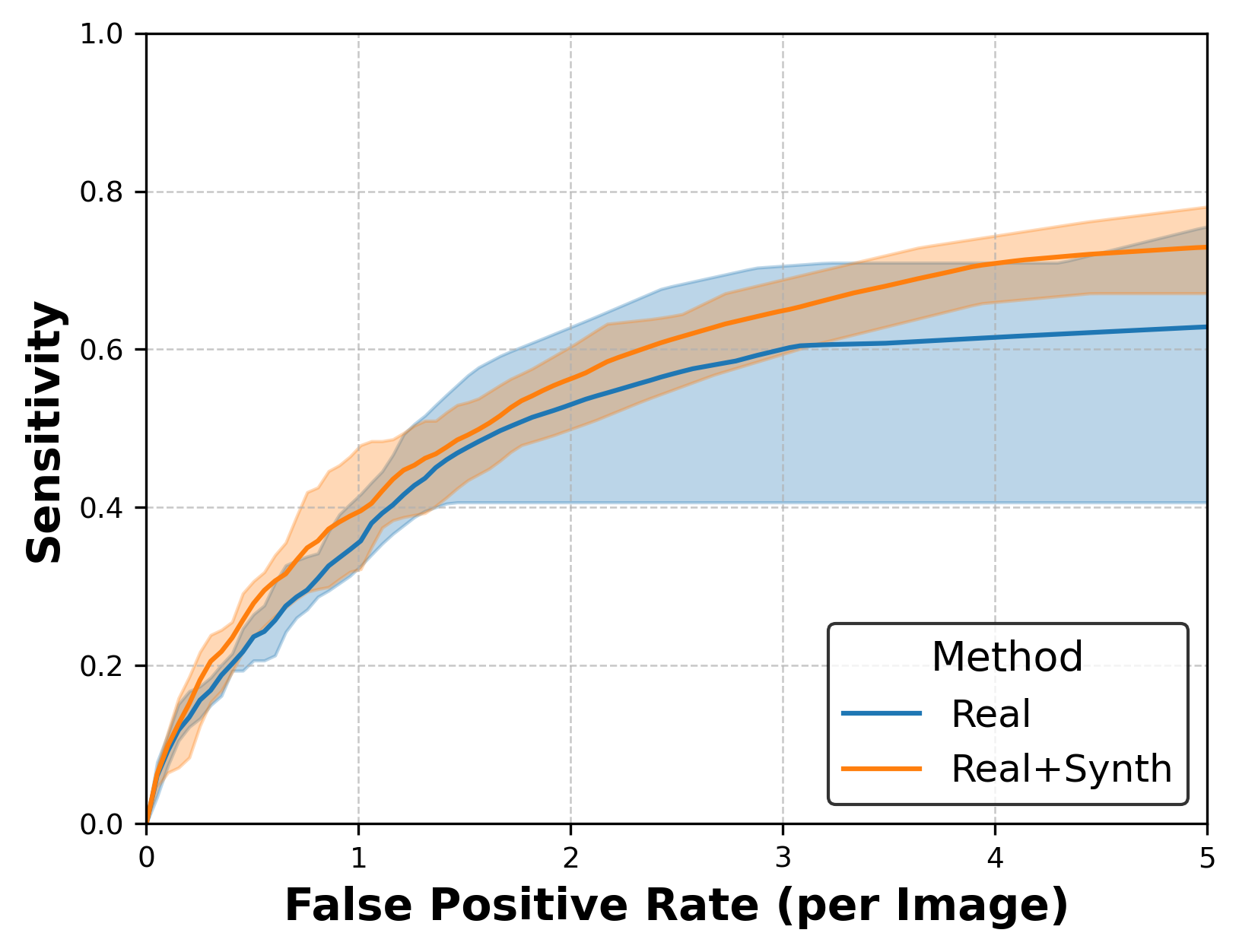}
    \label{fig:froc_real_synth_dbt_overall}
  \end{subfigure}  
  \begin{subfigure}{0.48\linewidth}
    \caption{DM Breast Density Subgroups}
    \includegraphics[width=0.9\linewidth]{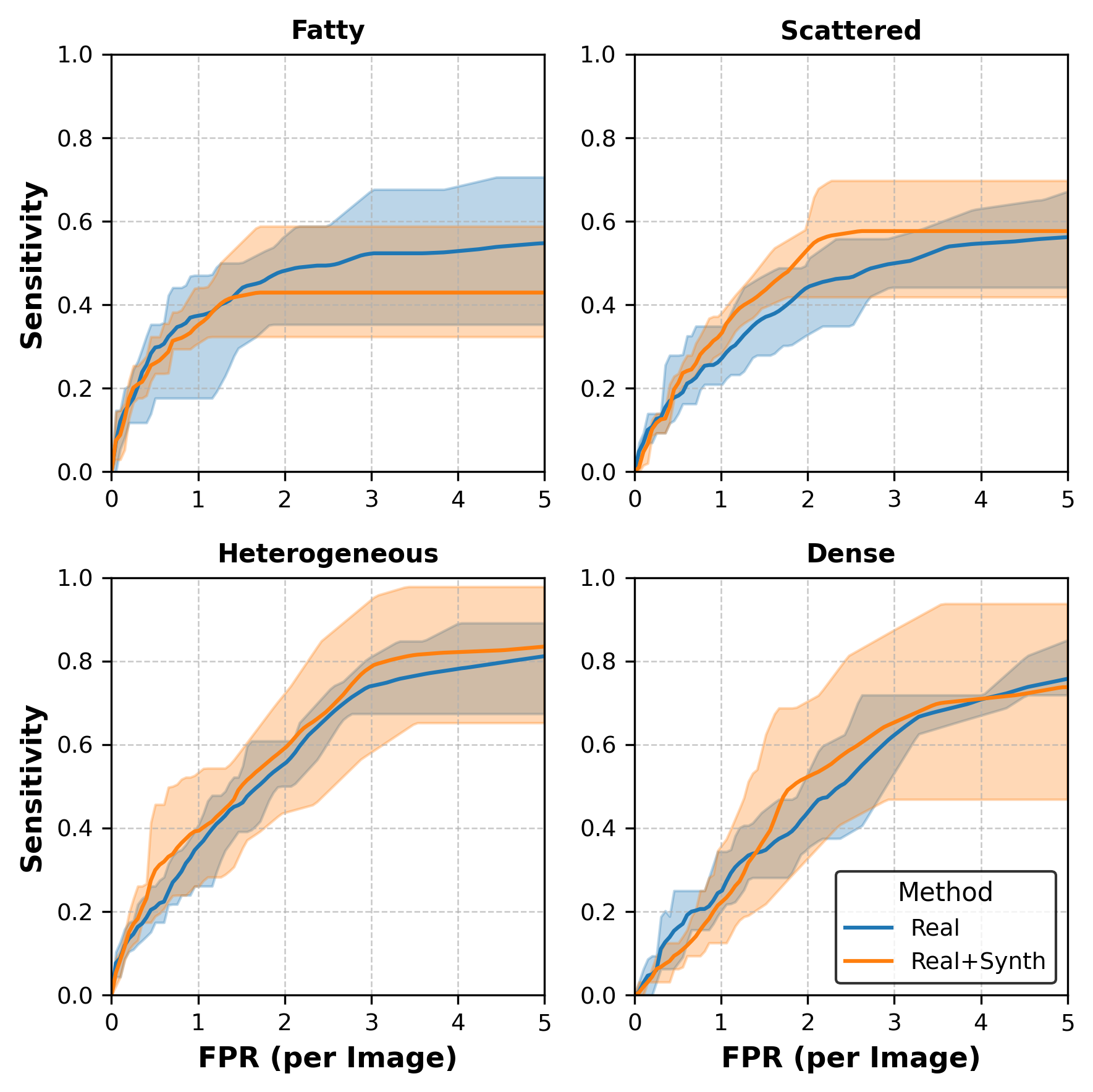}
    \label{fig:froc_real_synth_dm_subgroup}
  \end{subfigure}
  \begin{subfigure}{0.48\linewidth}
    \caption{DBT Breast Density Subgroups}
    \includegraphics[width=0.9\linewidth]{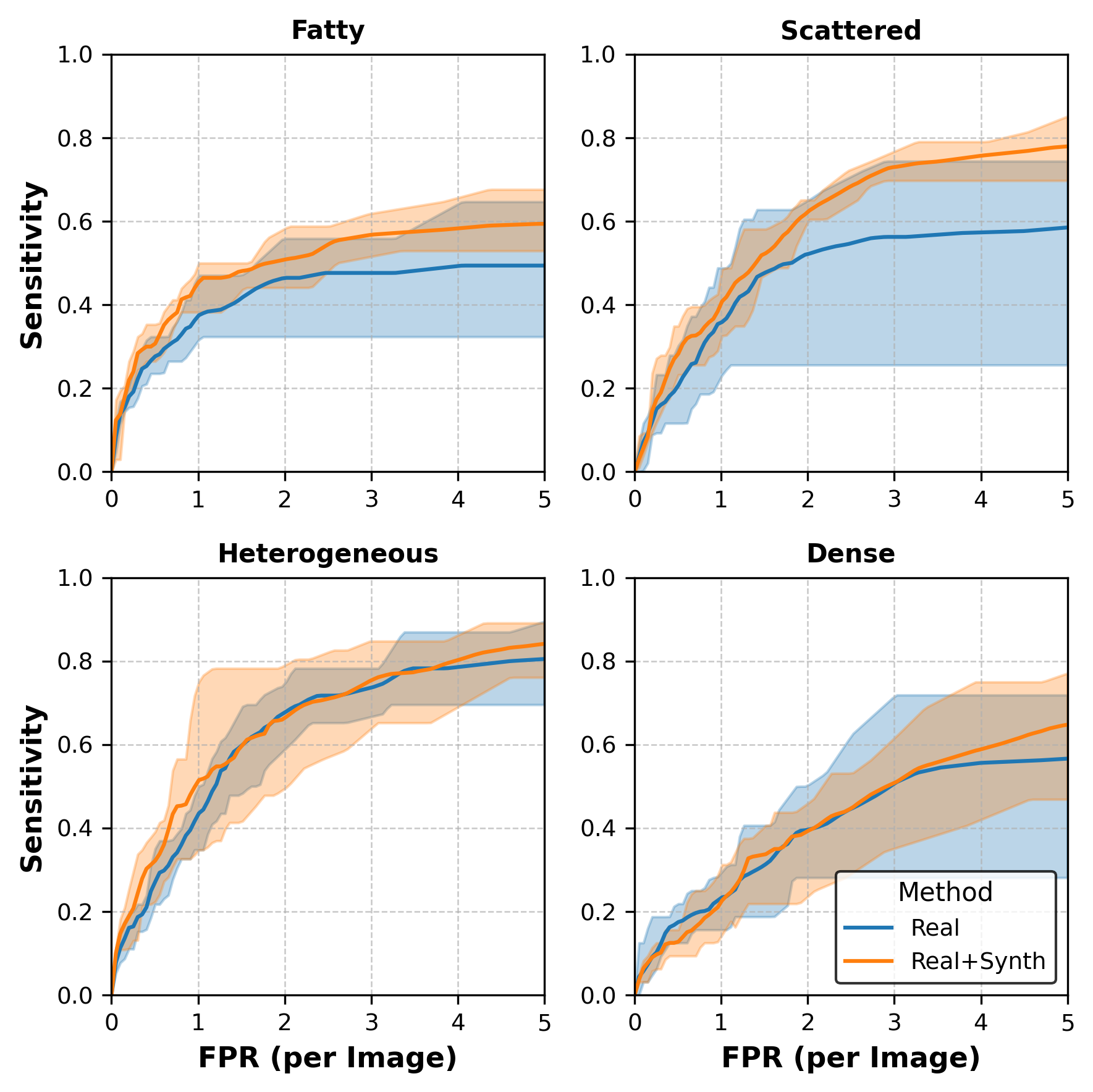}
    \label{fig:froc_real_synth_dbt_subgroup}
  \end{subfigure}
  \caption{Comparison of Digital Mammography (DM) and Digital Breast Tomosynthesis (DBT) Free-response ROC curve on real test set for detection models trained with real (EMBED~\cite{jeong2023emory}) and synthetic (T-SYNTH) data. Solid \textcolor{blue}{blue} and \textcolor{orange}{orange} lines represent average sensitivity across trials; spread captures minimum and maximum sensitivity.}
  \label{fig:froc_real_synth}

\end{figure*}

If we compare detection results on DM and DBT (see Fig.~\ref{fig:froc_real_synth}), we find that the corresponding results from DM, while having slightly lower and more variant performance, are generally consistent with the trends in DBT. Synthetic data appears to generate a slightly larger performance improvement in DBT compared to DM (e.g., compare scattered and dense subgroups). 

\subsection{Table of Public Breast Image Datasets} \label{sec:mammo_datasets}
Finally, we provide a summary of existing public breast image datasets (both patient and synthetic) in Tab.~\ref{tab:mammo_datasets} for additional information. 
\begin{table*}[htb]
    \centering
    \resizebox{\linewidth}{!}{
    \begin{threeparttable}
    \begin{tabular}{|l|c|c|c|c|c|c|c|c|c|}
        \hline
        \multicolumn{10}{|c|}{\textbf{Patient datasets}} \\
        \hline
        {\textbf{Dataset}} & {\textbf{DM}} & {\textbf{DBT}} & {\textbf{Bi-RADS}} & {\textbf{Density}} & {\textbf{\# cases}} & {\textbf{\# images}} & {\textbf{Image categories}} & {\textbf{Pop}} & {\textbf{Seg. Annotation}}  \\             
        \hline 
        BCS-DBT~\citep{buda2020detection} & No & Yes & Yes (5) & No & 5,060 & 22,032 & 
        \begin{tabular}{@{}c@{}}
        Cancer \\
        benign \\
        actionable \\
        normal
        \end{tabular}
        & USA & 
        \begin{tabular}{@{}l@{}}
        336 mass bboxes  \\
        99 arch.distortions bboxes
        \end{tabular}
        \\
        \hline
        ADMANI~\citep{frazer2022admani} & Yes & No & No & No & 629,863 & 4,411,263\tnote{a} &
        \begin{tabular}{@{}l@{}}
        Normal \\ recall 
        \end{tabular}
        & 
        \begin{tabular}{@{}l@{}}
        Several\\ nations 
        \end{tabular}
        & 
        \begin{tabular}{@{}l@{}}
        Annotations provided \\
        \textbf{where available} \\
        \end{tabular}
        \\
        \hline
        EMBED~\citep{jeong2023emory} & Yes & Yes & Yes (6) & Yes (4)\tnote{m} & 116,000 & 3,383,659\tnote{b} & 
        \begin{tabular}{@{}c@{}}
        Invasive cancer \\ non-invasive cancer \\ high risk
        \end{tabular}
        & USA\tnote{c} & 
        \begin{tabular}{@{}c@{}}
        40,000 lesion-level ROIs \\ with annotations \tnote{h}
        \end{tabular}\\
        \hline
        CMMD~\citep{cui2021chinese} & Yes & No & No & No & 1,775 & 3,712 & Benign, malignant & China & None \\
        \hline
        INBreast~\citep{moreira2012inbreast} & Yes & No & Yes & Yes (3) & 115 & 410 & Benign, malignant, normal & Portugal & 105 annotated images \\
        \hline
        OPTIMAM~\citep{Optimam} & Yes & No & No & No & 172,282 & 3,072,878 & Normal, interval cancers, benign, malignant & UK & 7,143 bboxes of lesions \\ 
        \hline
        VinDR~\citep{vindr} & Yes & No & Yes (6) & Yes (4) \tnote{z} & 5,000 & 20,000 & Normal, interval cancers, benign, malignant & UK & 20,000 bboxes of lesions \\ 
        \hline
        CBIS-DDSM~\citep{CBIS-DDSM} & Yes & No & Yes (5) & Yes (4) \tnote{z} & 6,775 & 10,239 & normal, benign, malignant & USA & 1,644 ROIs of masses \tnote{h} \\ 
        \hline
        BCDR~\citep{BCDR} & Yes & No & Yes (5) & Yes (3) \tnote{l} & 1,010 & 3,703 & 
        \begin{tabular}{@{}c@{}}
        masses \\ microcalcifications \\
        calcifications \\ stromal distortions \\ architectural distortions \\ axillary adenopathy
        \end{tabular}
         & USA & 2,335 lesion segmentations \\ 
        \hline
        DMID~\citep{DMID} & Yes & No & Yes (6) & Yes (3) \tnote{l} & 510 & 510 & 
        \begin{tabular}{@{}c@{}}
        mass-benign \\
        mass-malignant \\
        calcifications \\
        architectural distortion \\
        asymmetry
        \end{tabular} & India & 510 pixel-level annotations (masks) of masses \\ 
        \hline
        CSAW~\citep{CSAW} & Yes & No & No & No & 499,807 & $\approx$2,000,000 & cancerous, healthy & Sweden & 1891 tumor pixel-level annotations \\ 
        \hline
        KAU-BCMD~\citep{KAU-BCMD} & Yes & No & Yes (5) & Yes (4) \tnote{z} & 1,416 & 5,662 & normal, benign, and malignant & Saudi Arabia & segmentations and bboxes for 5,662 masses \\ 
        \hline
        Sheba~\citep{sheba} & Yes & No & No & No & 19 & 42 & isolated, clustered microcalcifications & Korea & 42 masks of microcalcifications \\ 
        \hline
        UC-Davis~\citep{Davis} & No & Yes & No & No & 150 & 150 & N/A & USA & 150 pixel-level lesion annotations\tnote{j} \\ 
        \hline
        MIAS Mammography~\citep{MIAS} & Yes & No & No & Yes (3)\tnote{l} & 416 & 416 & normal, benign, malignant & USA & 
        \begin{tabular}{@{}l@{}}
        416 ROIs of masses \tnote{i}
        \end{tabular}
        \\ 
        \hline
        NYU [Not publicly available] & Yes & No & Yes (3) & Yes (5) \tnote{z} & 229,426 & 1,001,093 & cancerous, healthy & USA & 8,080 pixel-level mass segmentations \\ 
        \hline
        \hline
        \multicolumn{10}{|c|}{\textbf{Synthetic datasets}}\\
        \hline
        {\textbf{Dataset}} & {\textbf{DM}} & {\textbf{DBT}} & {\textbf{Bi-RADS}} & {\textbf{Density}} & {\textbf{\# cases}} & {\textbf{\# images}} & {\textbf{Image categories}} & {\textbf{Pop}} & {\textbf{Seg. Annotation}}  \\
        \hline
        Sarno~\citep{Sarno2021_digitaldataset} & Yes & Yes & -- & -- & -- & 150\tnote{d} & Normal & -- & No \\
        \hline
        VICTRE~\citep{Badano2018victre} & Yes & Yes & -- & -- & -- & 2986 & Negative, positive cohort & -- & Yes\tnote{e}  \\
        \hline        
        CSAW~\citep{csaw_m} & Yes & No & -- & -- & -- & 100k & high, medium, low masking levels & -- & Yes\tnote{n} \\ 
        \hline 
        M-SYNTH~\citep{sizikova2024knowledge} & Yes & No\tnote{g} & -- & -- & -- & 45,000 & Negative, positive cohort & -- & Yes\tnote{f} \\ 
        \hline
        T-SYNTH (Ours) & Yes & Yes\tnote{g} & -- & -- & -- & 9,000 & Negative, positive cohort & -- & Yes\tnote{f} \\
        \hline
    \end{tabular}
    \begin{tablenotes}
        \footnotesize
        \item[a] subset available for the RSNA Cancer Detection AI challenge.
        \item[b] 20\% available via AWS; contains annotated lesions.
        \item[c] equal representation of African American and White
        \item[d] 150 uncompressed, 60 compressed images
        \item[e] four breast densities, same lesions across all positive cohort
        \item[f] 3 lesion densities, 3 lesion sizes, 4 breast densities, 5 different doses
        \item[g] A corresponding DBT image dataset will be provided in a future release of the dataset.
        \item[h] location of centroid + radius. Saved as a white circle directly onto a copy of the original mammogram, generating a screen save image.
        \item[i] location of centroid + radius.
        \item[j] contains computational digital breast phantoms. Each image voxel was classified in one out of the four main materials presented in the field of view: fibroglandular tissue, adipose tissue, skin tissue, and air
        \item[l] fatty, fatty-glandular, dense-glandular
        \item[m] high density, isodense, low density, fat containing
        \item[n] diffusion-based generator
        \item[z] almost entirely fatty (1), scattered areas of fibroglandular density (2), heterogeneously dense (3), extremely dense (4), unknown (5)
    \end{tablenotes}
    \end{threeparttable}
    }
    
    \caption{Summary of existing breast imaging datasets. The proposed T-SYNTH dataset is the largest synthetic paired DM-DBT dataset available.}
    \label{tab:mammo_datasets}
\end{table*}

\subsection{Diffusion Experiments}
The diffusion model is run using an auto-inpainting pipeline from Hugging Face, where an image without a lesion, a circular mask, and a prompt (``draw a lesion'') are presented to the model, which outputs a breast image with the lesion drawn in the specified location. This approach allowed us to generate both images and bounding boxes for each generated lesion. 
\begin{itemize}
\item \textbf{T-SYNTH}: positive (300 from T-SYNTH + 100 from EMBED), negative (400 from EMBED)
\item \textbf{EMBED baseline}: positive (400 from EMBED), negative (400 from EMBED)
\item \textbf{Subset EMBED baseline}: positive (100 from EMBED), negative (400 from EMBED)
\item \textbf{Diffusion}: positive (300 from un-finetuned diffusion, 100 from EMBED), negative (400 from EMBED)
\item \textbf{Finetuned Diffusion}: positive (300 from finetuned diffusion, 100 from EMBED), negative (400 from EMBED)
\end{itemize}

\bibliography{references}

\end{document}